\documentclass[10pt,twocolumn,letterpaper]{article}

\usepackage[pagenumbers]{cvpr} % To force page numbers, e.g. for an arXiv version

\usepackage[dvipsnames]{xcolor}

\usepackage{multirow}
\usepackage{booktabs}
\definecolor{cvprblue}{rgb}{0.21,0.49,0.74}
\usepackage[pagebackref,breaklinks,colorlinks,citecolor=cvprblue]{hyperref}

\title{Parallax-tolerant Image Stitching via Segmentation-guided Multi-homography Warping}

\author{Tianli Liao, Lei Li and Guangen Liu\\
College of Information Science and Engineering, Henan University of Technology\\
{\tt\small \{tianli.liao, leili, lgendd\_99\}@haut.edu.cn}
\and
Ce Wang\\
Hong Kong University of Science and Technology\\
{\tt\small fogever@icloud.com}
\and
Nan Li\\
School of Mathematical Sciences, Shenzhen University\\
{\tt\small nan.li@szu.edu.cn}
}

\begin{document}

\maketitle

\begin{abstract}
	Large parallax between images is an intractable issue in image stitching. Various warping-based methods are proposed to address it, yet the results are unsatisfactory. In this paper, we propose a novel image stitching method using multi-homography warping guided by image segmentation. Specifically, we leverage the Segment Anything Model to segment the target image into numerous contents and partition the feature points into multiple subsets via the energy-based multi-homography fitting algorithm. The multiple subsets of feature points are used to calculate the corresponding multiple homographies. For each segmented content in the overlapping region, we select its best-fitting homography with the lowest photometric error. For each segmented content in the non-overlapping region, we calculate a weighted combination of the linearized homographies. Finally, the target image is warped via the best-fitting homographies to align with the reference image, and the final panorama is generated via linear blending. Comprehensive experimental results on the public datasets demonstrate that our method provides the best alignment accuracy by a large margin, compared with the state-of-the-art methods. The source code is available at \url{https://github.com/tlliao/multi-homo-warp}.
	%\keywords{Image stitching \and image warping \and homography warp \and large parallax}
\end{abstract}

\section{Introduction}
\label{sec:intro}

\begin{figure*}[t]
	\centering
	\subfloat[Input images]{
		\includegraphics[height=0.107\textheight]{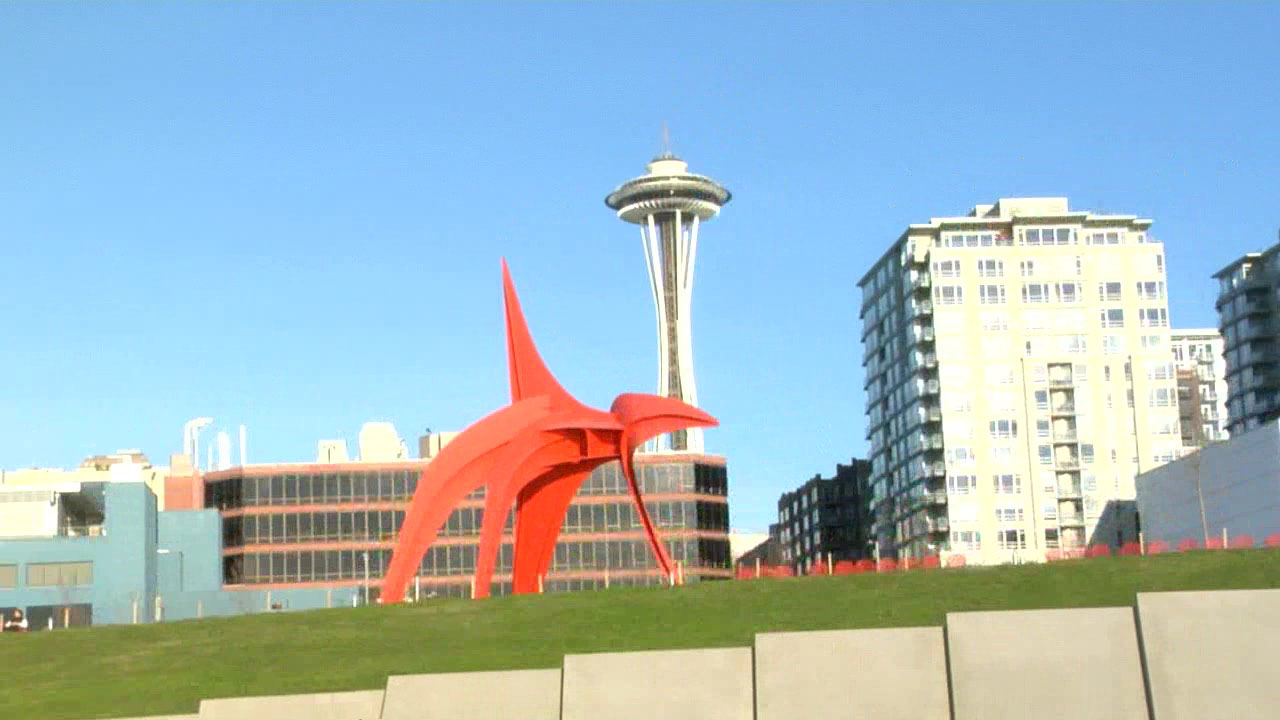}
		\includegraphics[height=0.107\textheight]{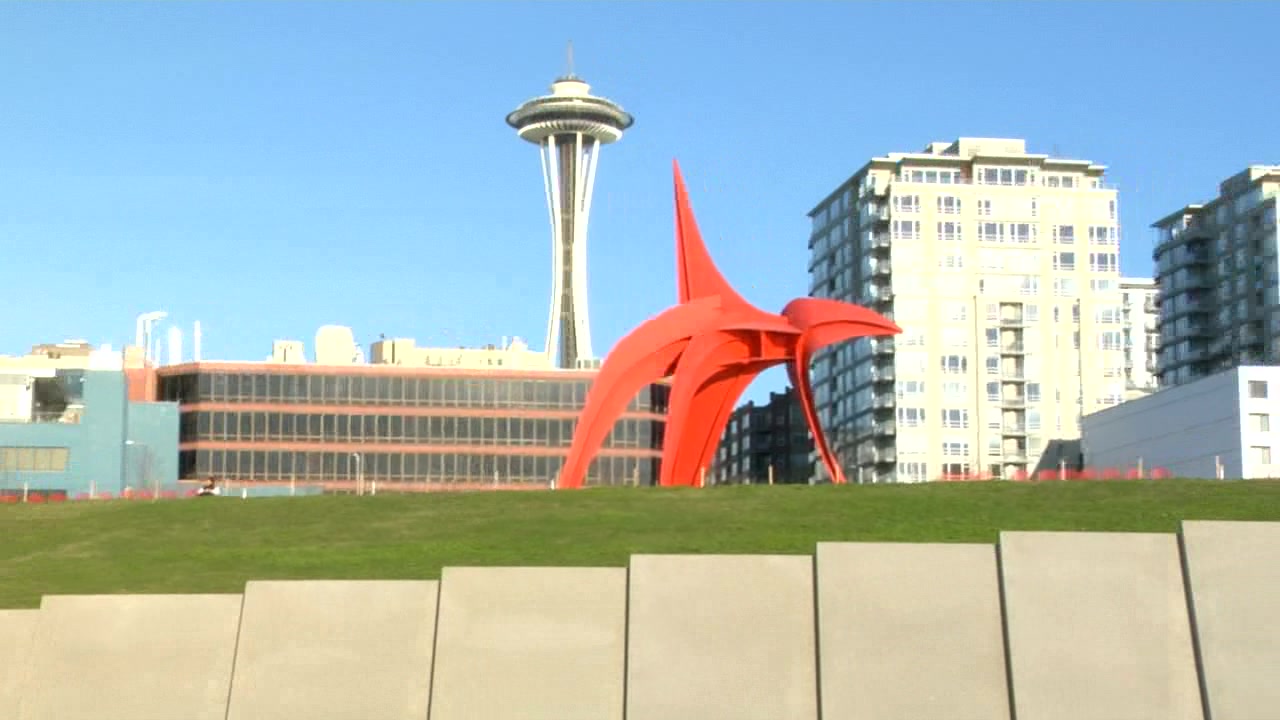}}
	\subfloat[SAM result of target image]{
		\includegraphics[height=0.107\textheight]{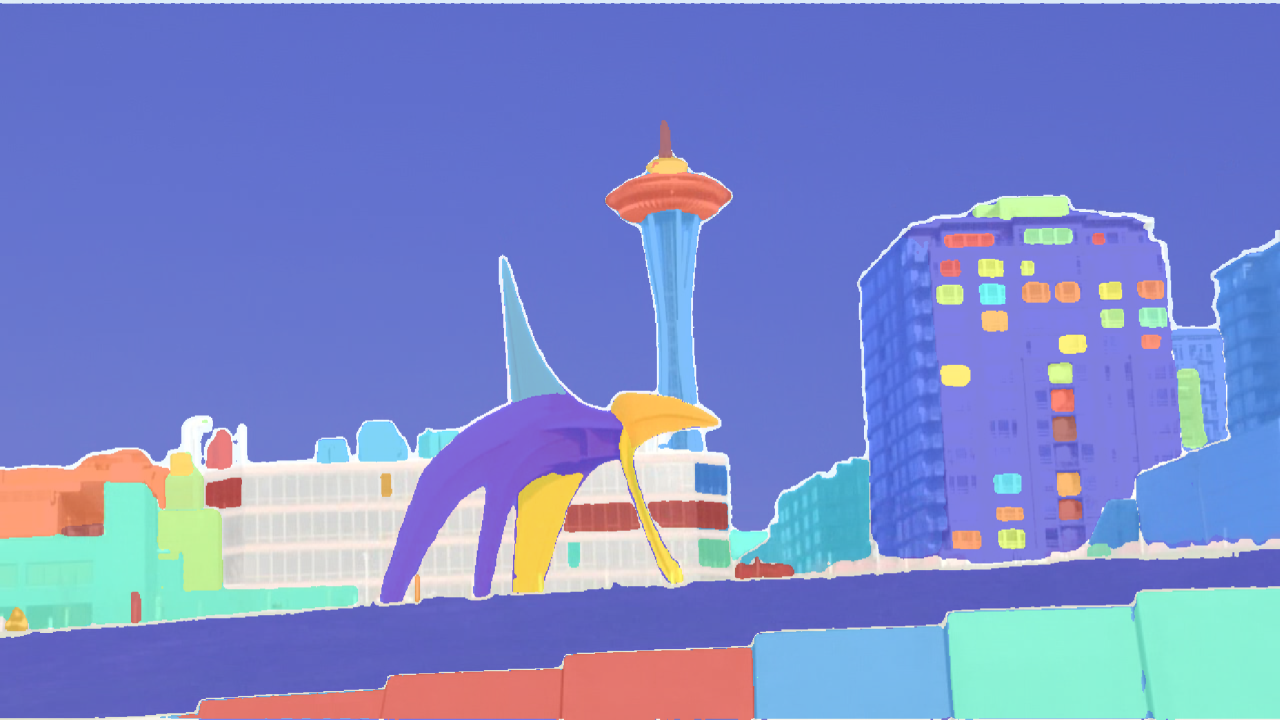}}
	\subfloat[Ours]{
		\includegraphics[height=0.107\textheight]{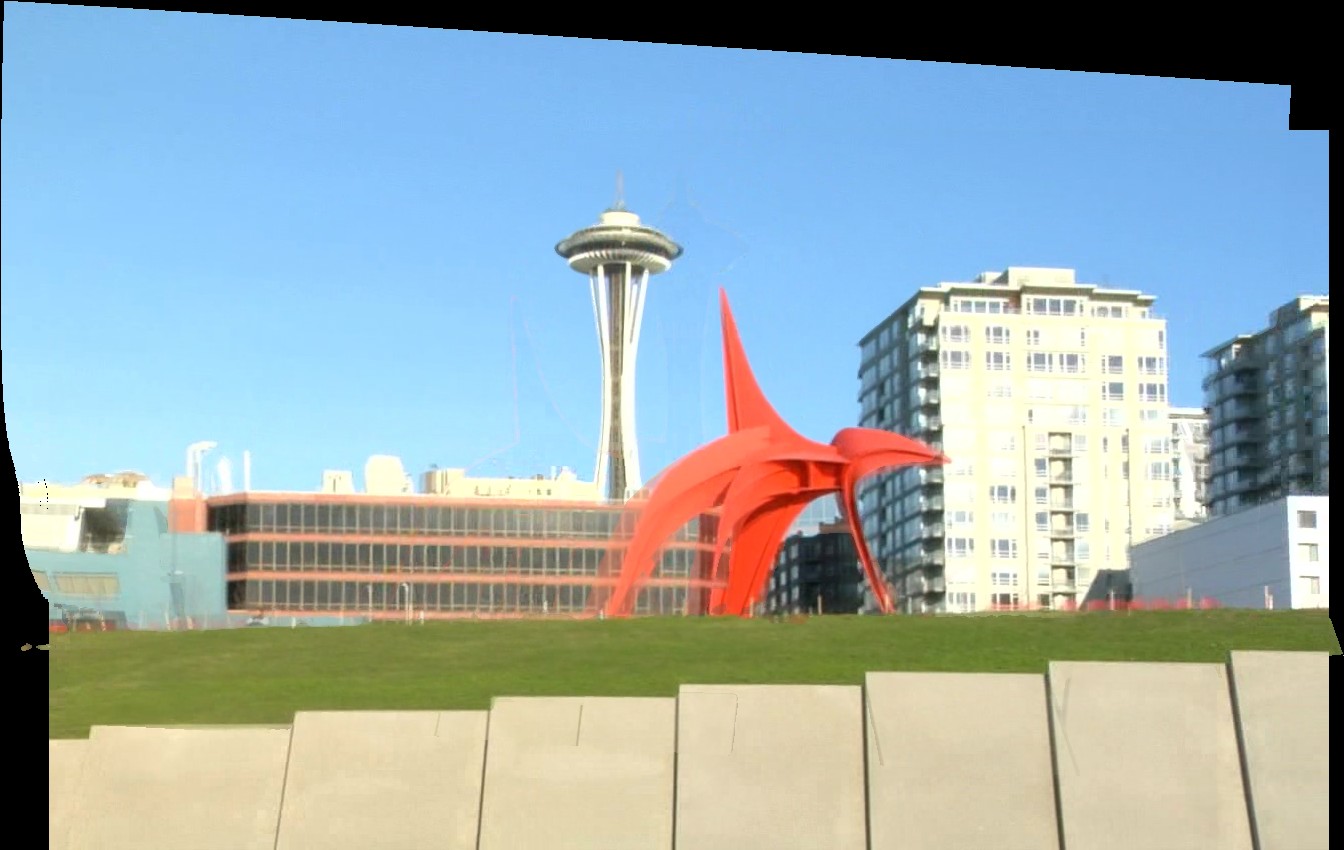}}\\	
	\subfloat[Baseline]{
		\includegraphics[height=0.11\textheight]{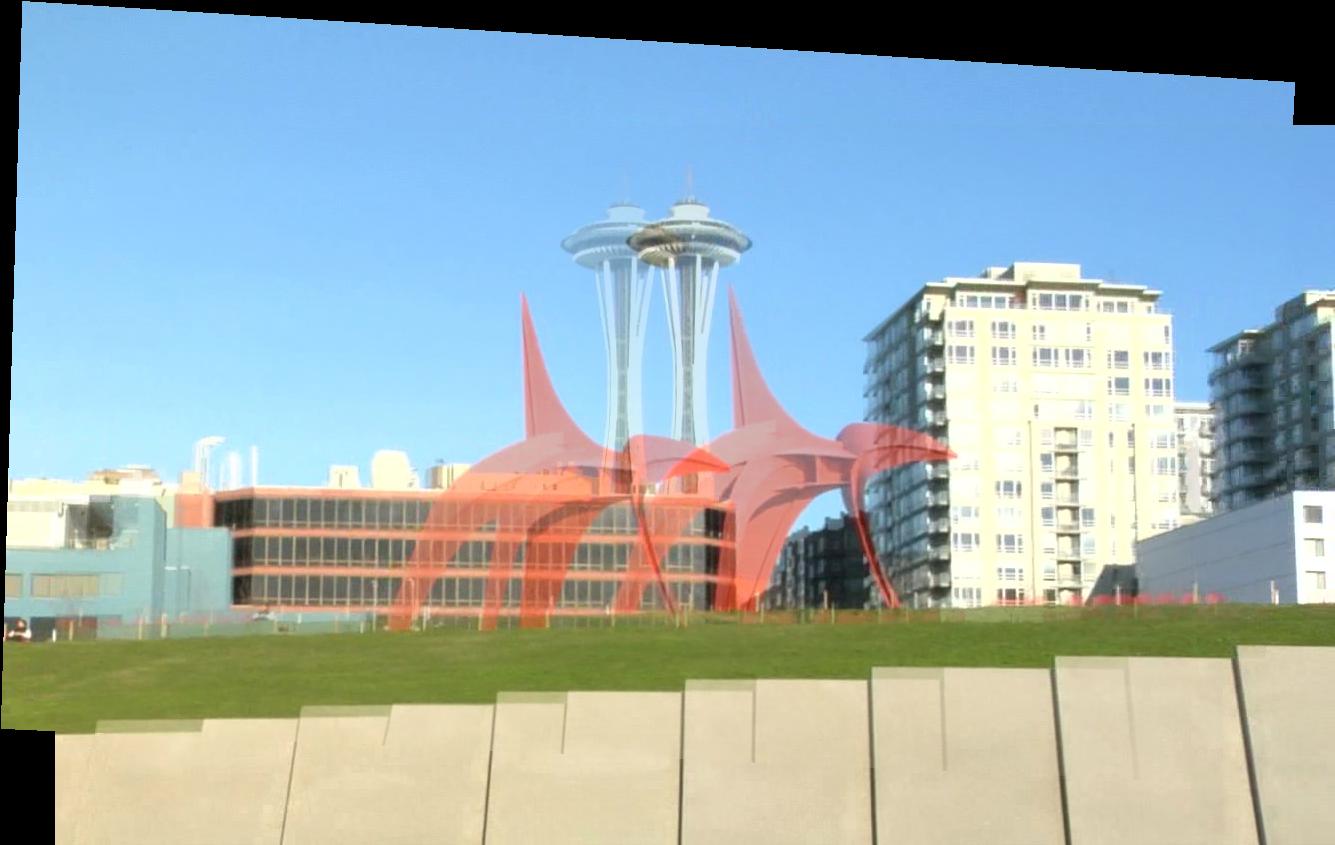}}
	\subfloat[APAP \cite{zaragoza2014projective}]{
		\includegraphics[height=0.11\textheight]{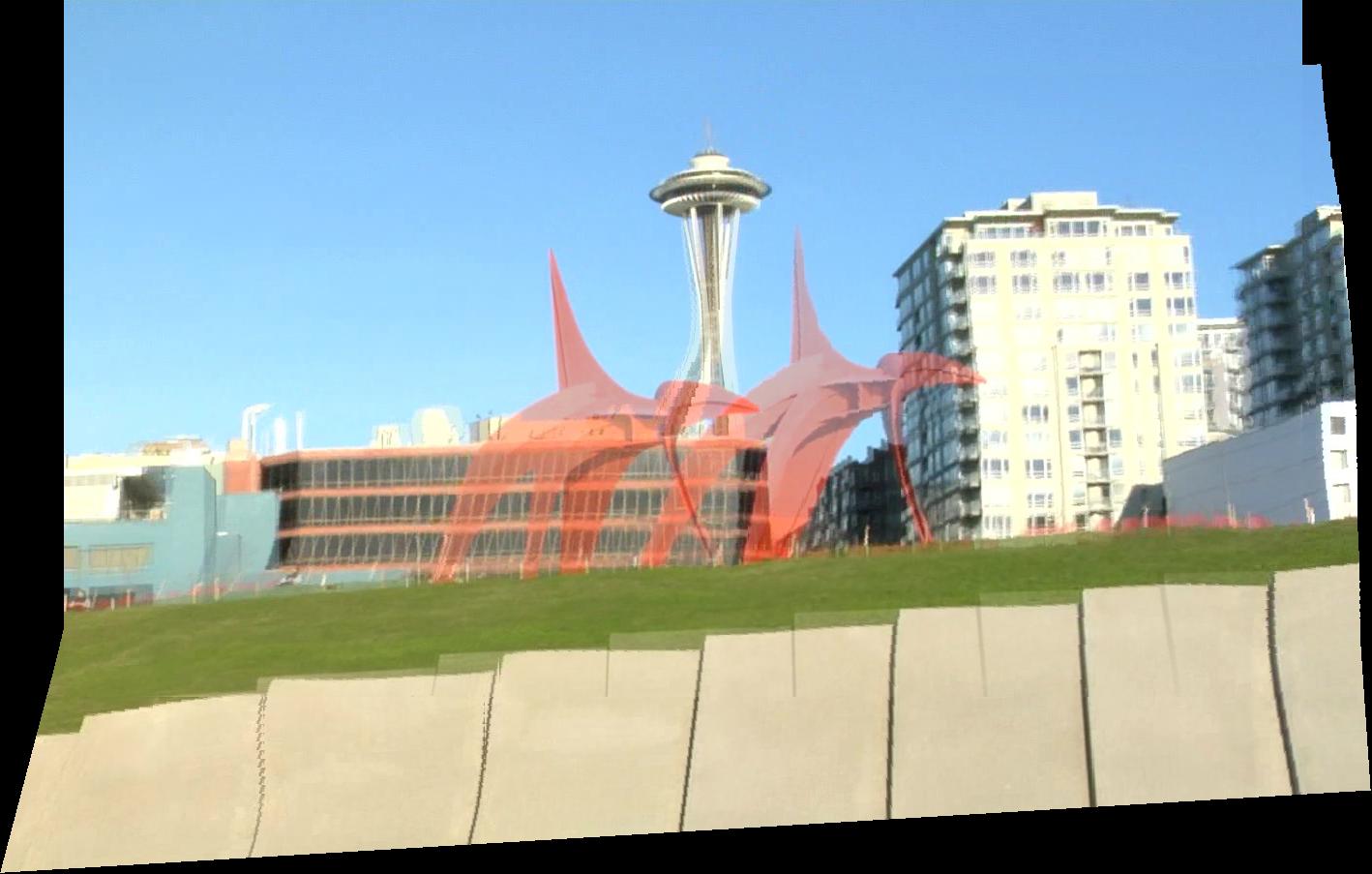}}
	\subfloat[REW \cite{li2018parallax}]{
		\includegraphics[height=0.11\textheight]{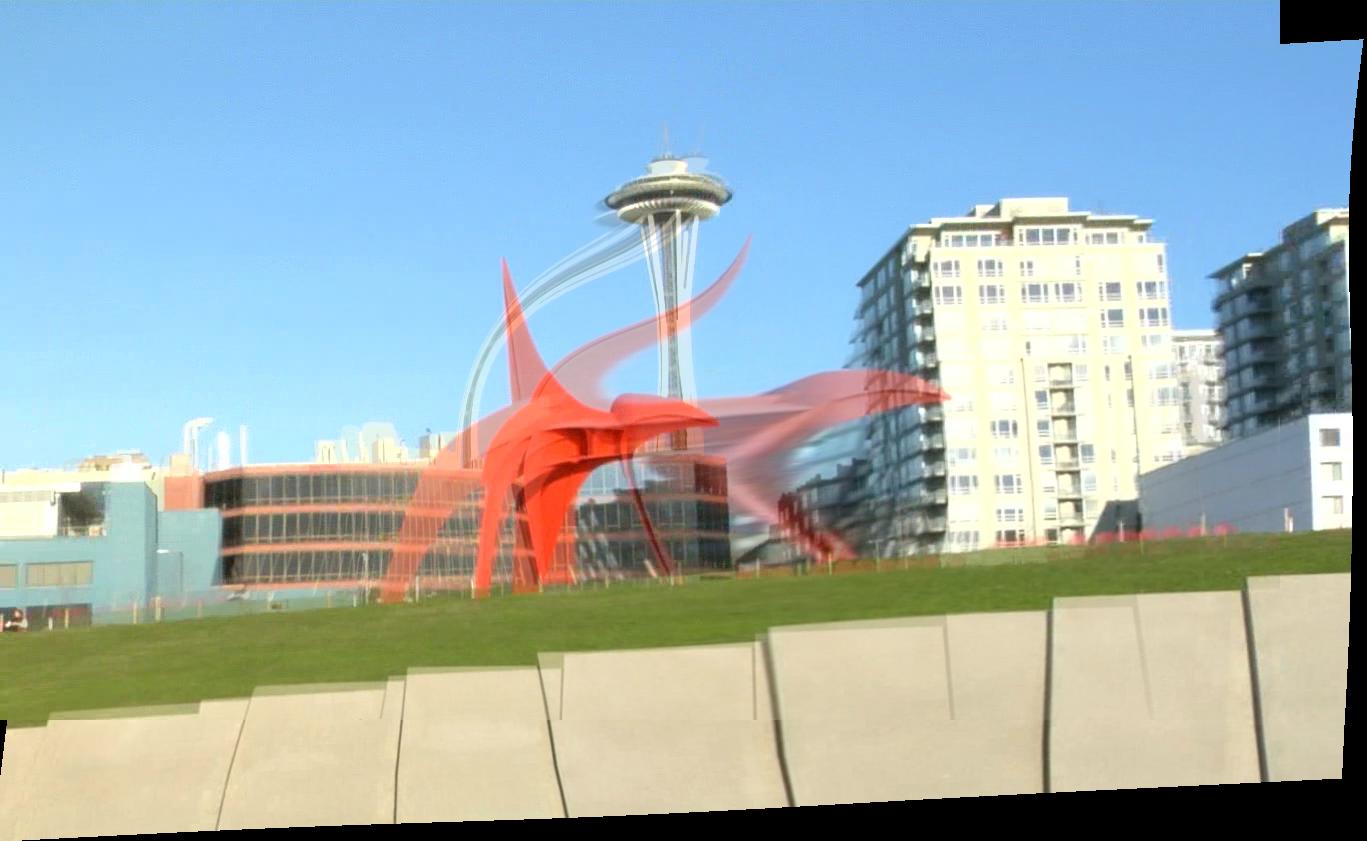}}
	\subfloat[TFA \cite{li2019local}]{
		\includegraphics[height=0.11\textheight]{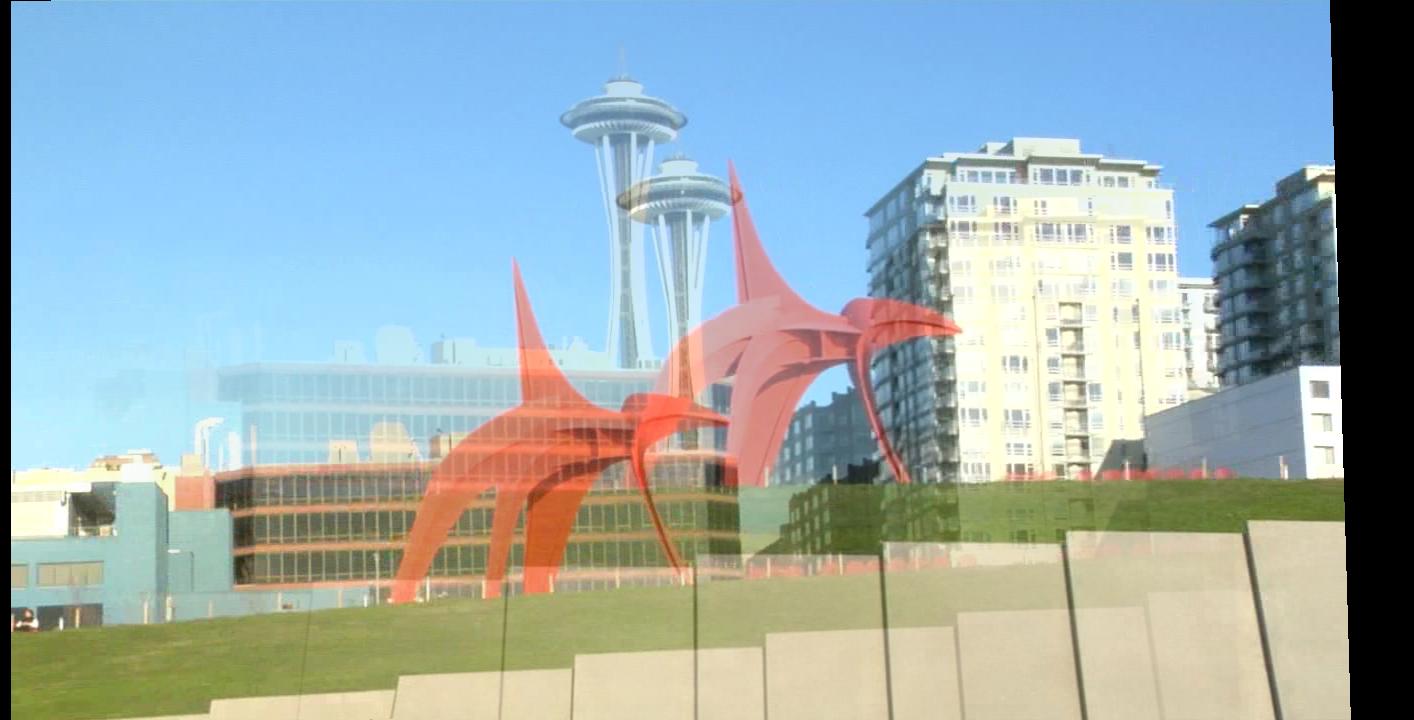}}\\
	\subfloat[GSP \cite{chen2016natural}]{
		\includegraphics[height=0.107\textheight]{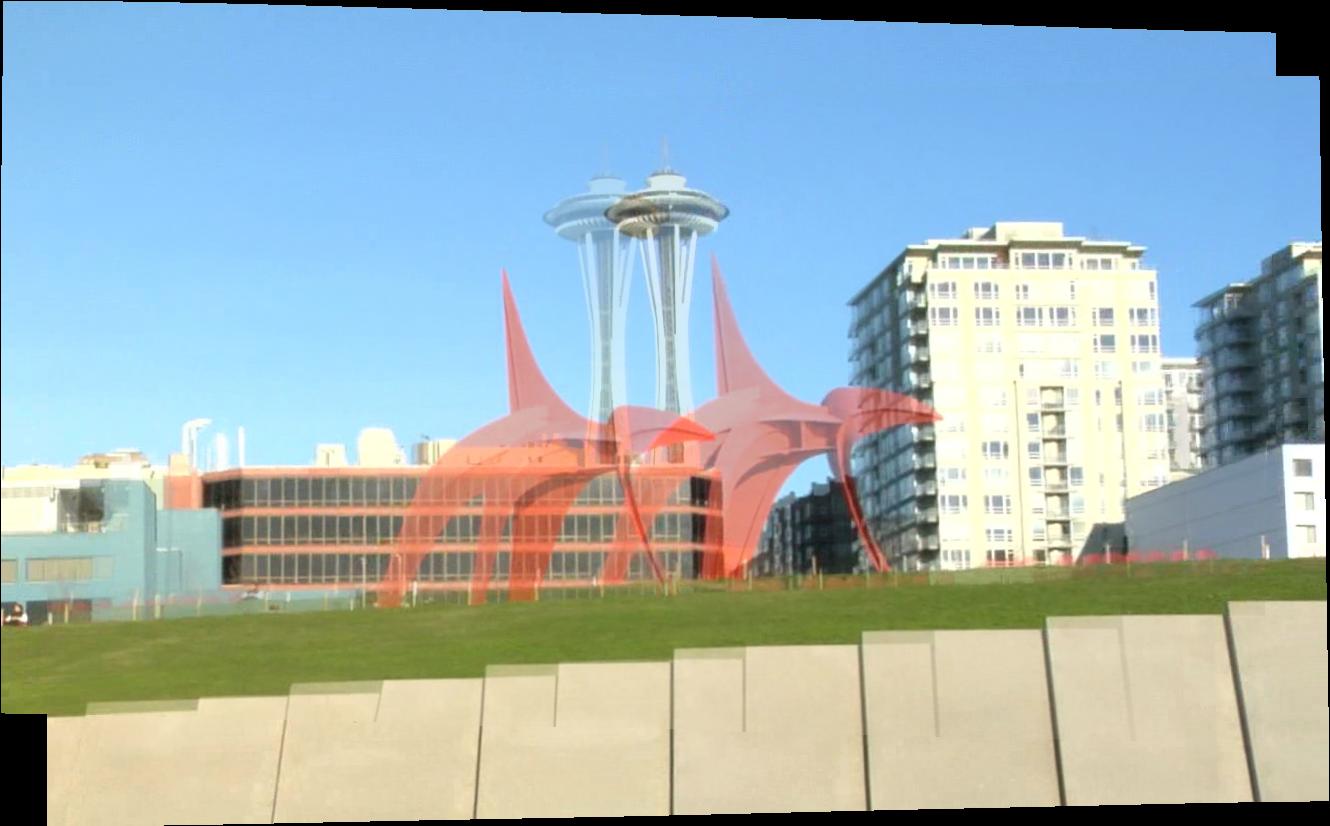}}	
	\subfloat[SPW \cite{liao2020Single}]{
		\includegraphics[height=0.107\textheight]{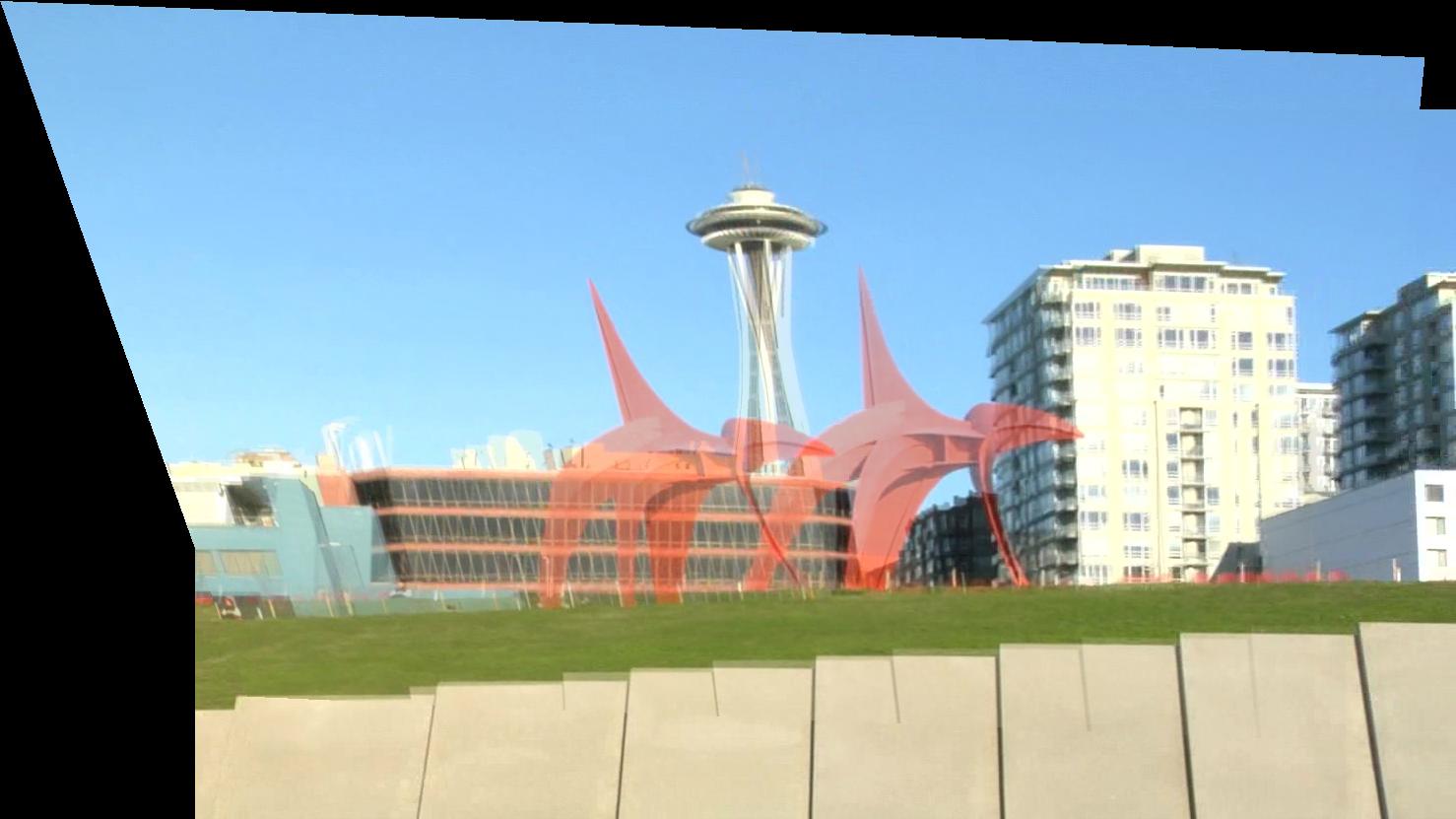}}
	\subfloat[LPC \cite{jia2021Leveraging}]{
		\includegraphics[height=0.107\textheight]{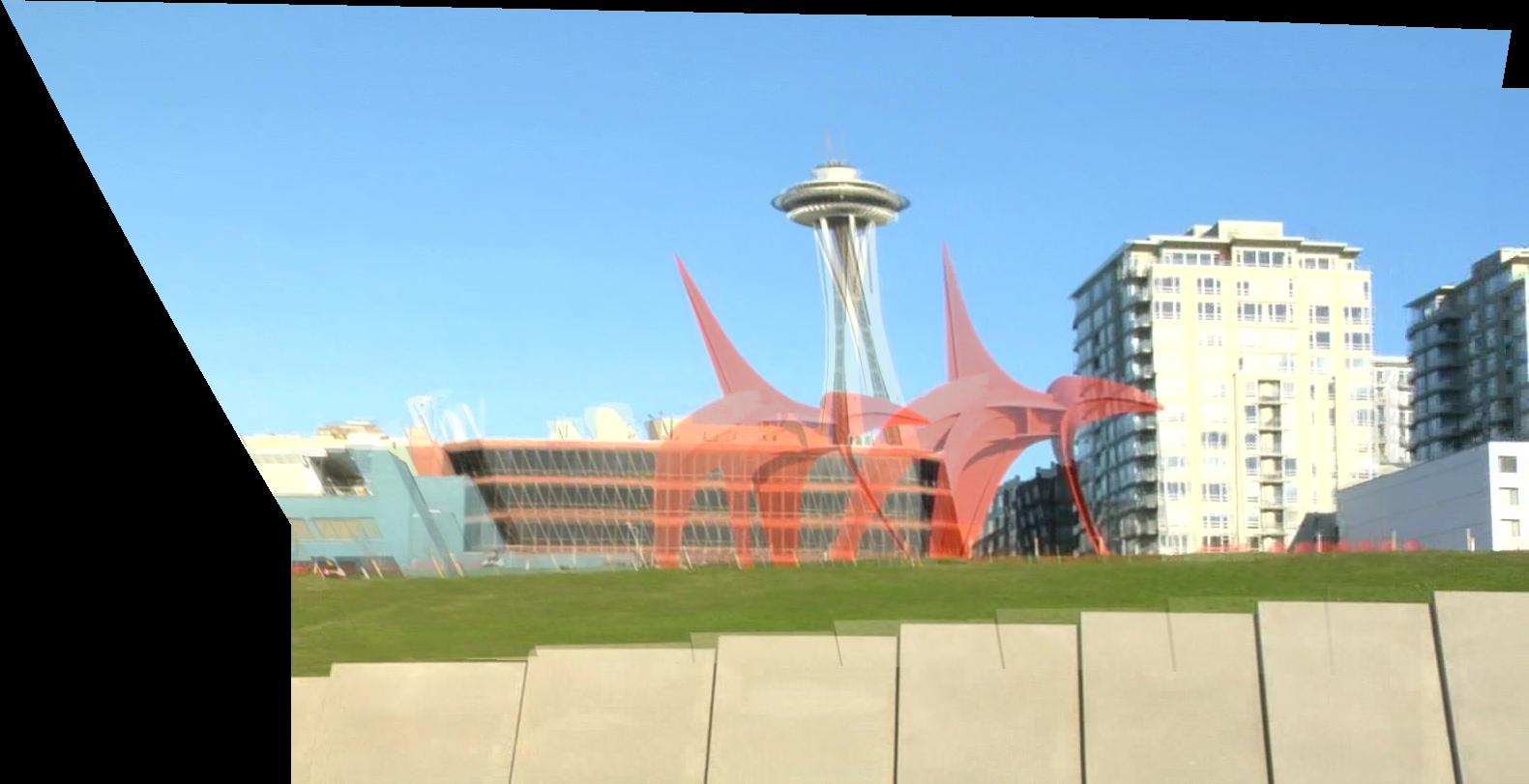}}	
	\subfloat[UDIS++ \cite{nie2023parallax}]{
		\includegraphics[height=0.107\textheight]{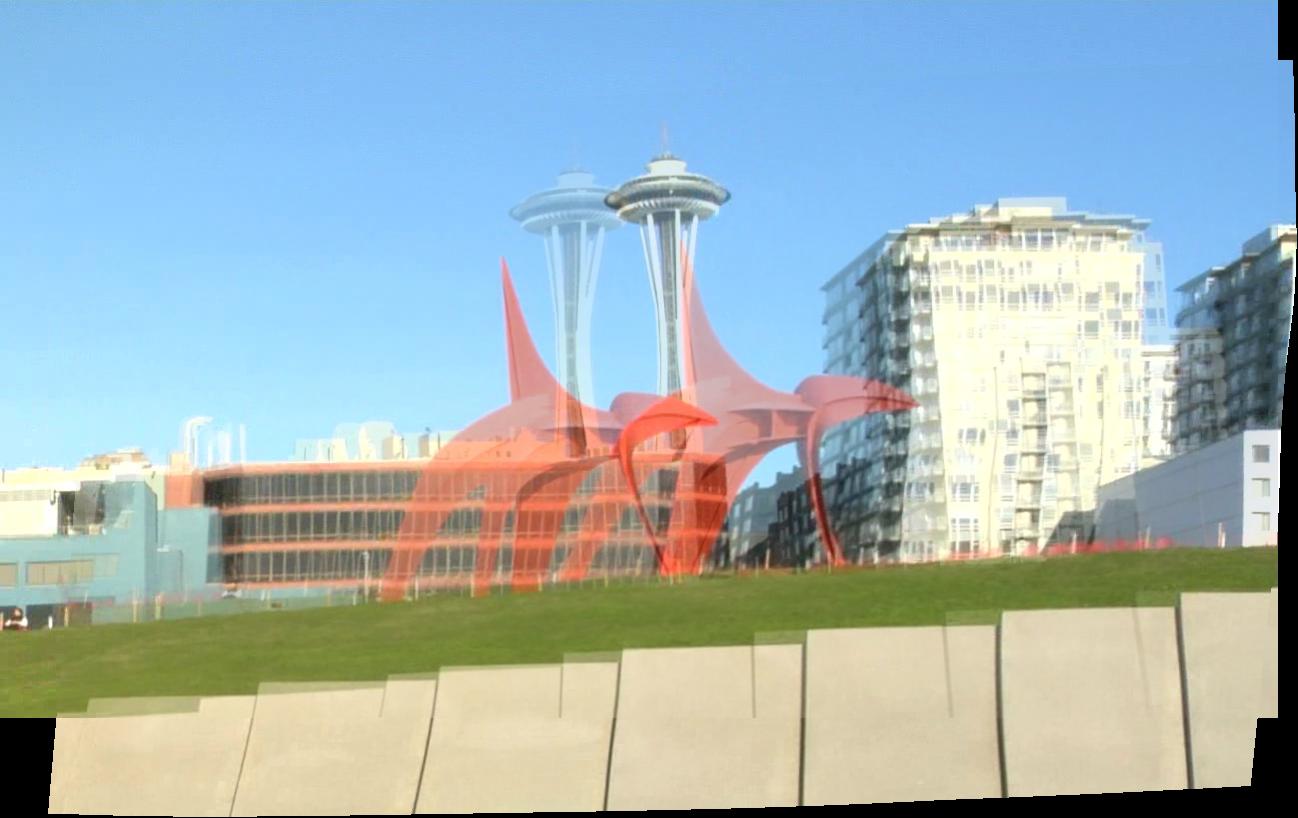}}\\			
	\caption{Stitching results comparison with various warping methods. All results are generated via simple linear blending. Obvious parallax can be seen from the tower and the red statue. (e)-(g) correspond to spatially-varying warps, (h)-(j) correspond to mesh-based warps, and (k) is the learning-based warp. Compared with other warps, our method achieves the best alignment quality.} % ANAP \cite{lin2015adaptive} fails on the image pair.
	\label{fig:1}
\end{figure*}	

Image stitching is a widely used technology in various computer vision applications and consumer photography. 
For captured images with a narrow field-of-view (FoV), the aim is to construct the panorama with a wider FoV scene.
%It aims to construct a wide field-of-view (FoV) scene from captured images with narrow FoV. 
Although, there are several academic and commercial stitching tools published, e.g., Adobe Photoshop Photomerge, Microsoft Image Composite Editor, and Autostitch \cite{Brown:2007}, image stitching with large parallax is still an intractable problem. Various stitching methods are proposed to handle it using different strategies.

Most of the existing methods follow a similar pipeline with two steps \cite{szeliski2006image}: 
Image warping and alignment are first followed by image composition and blending. Generally, the crucial image warping step determines the stitching quality, and the homography is a simple and commonly used warping model representing the planar transformation between the input scenes. 
However, it has the assumptions of the planar captured scenes or purely rotational camera motion.
Violating such assumptions leads to an undesirable parallax between images, making the homography fail to deal with. As the case of Fig.~\ref{fig:1}(d) shows, severe ghosting artifacts are generated in the stitching results.

To alleviate the artifacts caused by parallax, adaptive warping models are proposed using either spatially-varying warps \cite{gao2011constructing,chang2014shape,Lee_2020_CVPR,li2017quasi,li2019local,lin2015adaptive,liu2019shape,zaragoza2014projective,Zheng2019tmm,lin2022image} or mesh-based warps \cite{chang2014SpatiallyVarying,chen2016natural,zhang2016multi,xiang2018Image,liao2020Single,jia2021Leveraging,li2021image,zhang2021natural,chen2021Image,du2022Geometric}.
In particular, the spatially varying warps aim to create a spatially varying motion field in the image domain. The warps provide better alignment in the overlapping region and can be smoothly extrapolated to the non-overlapping region. To formulate the warping model as the energy minimization of the mesh deformation, mesh-based warps partition the images into regular meshes. 
% Mesh-based warps partition the images into regular meshes and formulate the warping model as the energy minimization of the mesh deformation. 
Then different energy functions are designed to maintain the smoothness of the warps, improving the alignment accuracy and preventing distortions. Designing the deep learning framework to address image stitching \cite{nie2023parallax} is another way to handle the parallax, while it still has the smoothness constraint. For images with larger parallax, the warps with smoothness constraints easily fail in aligning the foreground objects and background scenes simultaneously, as shown in Fig.~\ref{fig:1}(e-k). \textit{Intuitively, aligning the objects and scenes separately by different homographies and fusing them will improve the alignment.} Motivated by the remarkable segmentation performance of the segment anything model (SAM) \cite{kirillov2023segment}, as shown in Fig.~\ref{fig:1}(b), we propose the multi-homography warping method which achieves the best alignment accuracy, as shown in Fig.~\ref{fig:1}(c).

In this paper, we propose a stitching method via multi-homography warping for images with large parallax. Firstly, we segment the target image into numerous contents via SAM and detect feature points from input images; followed by partitioning them into multiple subsets via the energy-based multi-homography fitting algorithm to generate multiple homographies. Then, for each segmented content in the overlapping region, we select (label) its best-fitting homography with the lowest photometric error. For each content in the non-overlapping region, we calculate (label) a weighted combination of the linearized homographies. Finally, the target image is warped based on the homographies and aligned with the reference image. Experimental results show that our method can accurately align different image contents with large parallax and outperforms the state-of-the-art warping methods by a large margin. The main contributions of our work are 3-fold as follows:
\begin{itemize}
	\item We propose a stitching method to integrate the SAM into the warping model generation and it achieves the best alignment accuracy. 
	\item The multi-homography fitting algorithm we designed can provide better homography models than the iterative RANSAC, especially for images with large parallax. Our algorithm is more stable and robust to the outliers of the feature matches. 
	\item We propose a parameter-free forward and backward warping strategy to warp different contents in the target image via multiple homographies and generate well-aligned results with correct occlusions (holes).
	
\end{itemize}

The rest of this paper is organized as follows. Section \ref{sec:related} introduces the related work of image stitching. Section \ref{sec:method} gives a detailed description of our multi-homography warping method, including multi-homography fitting, multi-labeling, image warping and blending. Section \ref{sec:exp} demonstrates the experimental results. Section \ref{sec:conclusion} concludes the paper.

\section{Related Work}
\label{sec:related}

\begin{figure*}[t]
	\centering
	\includegraphics[width=0.8\textwidth]{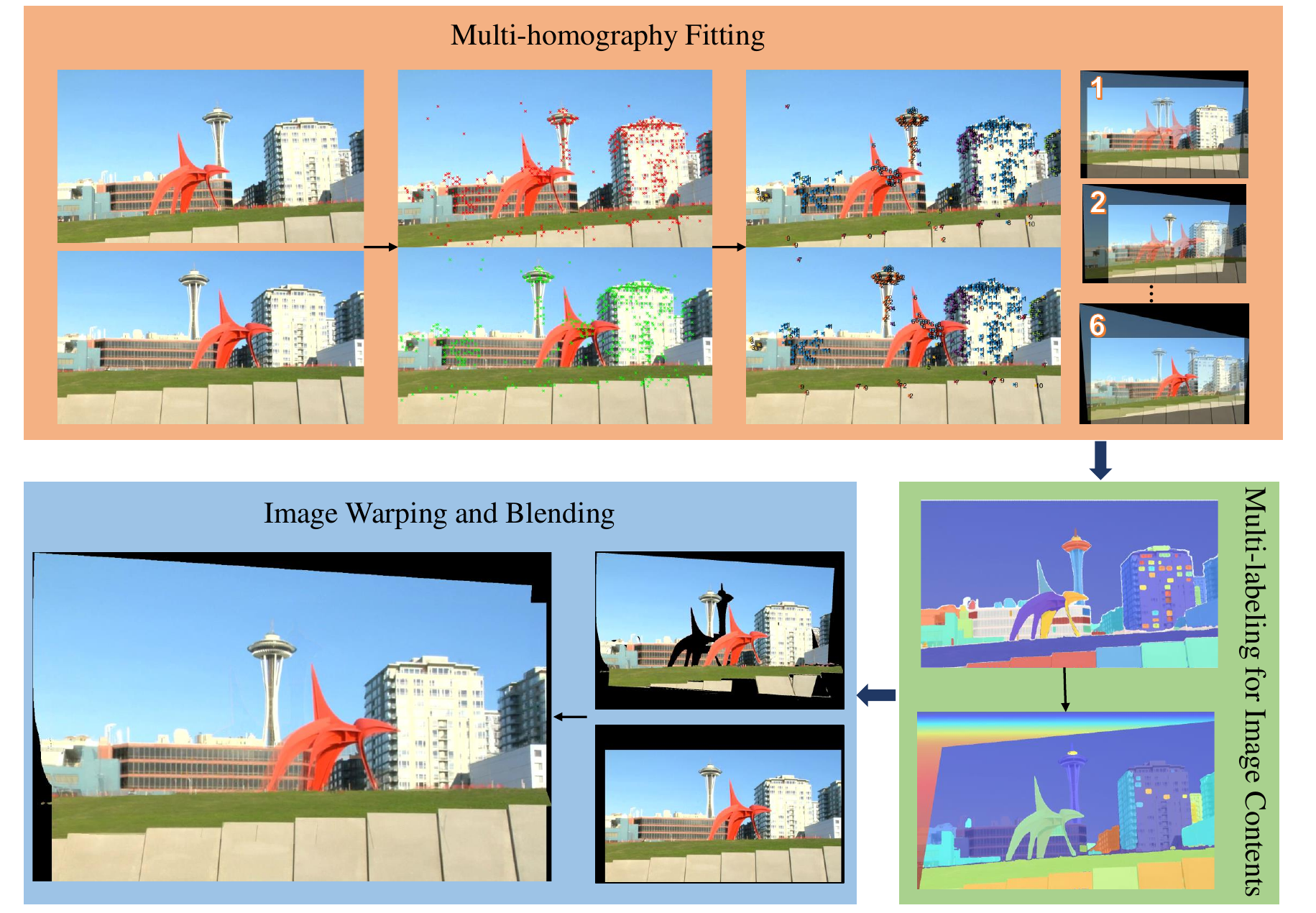}
	\caption{Pipeline of our multi-homography warping method.}
	\label{fig:pipeline}
\end{figure*}
%-------------------------------------------------------------------------
\subsection{Spatially-varying Warps}

Suppose a set of feature matches is given, some warping methods adopted piece-wise homographies as adaptive warping models where every local homography is determined via some weighting methods. Gao \emph{et al.} \cite{gao2011constructing} proposed
a dual-homography (DH) warping model, where two representative homographies for two dominant planes (distant plane and ground plane) are estimated. Then the spatially-varying motion field is calculated by interpolating the two homographies as basis motions. Zheng \emph{et al.} \cite{Zheng2019tmm} further developed DH to handle the overlapping scene consisting of multiple dominant planes.

Zaragoza \emph{et al.} \cite{zaragoza2014projective} proposed an as-projective-as-possible (APAP) warping model to handle parallax. It generates the smooth spatially-varying motion field via fitting weighted correspondences in feature matches instead of interpolating basis motion. For each quad mesh, it estimates a location-dependent homography by weighted direct linear transform. Li \emph{et al.} \cite{li2019local} and Lee \emph{et al.} \cite{Lee_2020_CVPR} improved the APAP model by estimating a weighed homography for each triangle and superpixel, respectively. Li \emph{et al.} \cite{li2018parallax} proposed a robust elastic warping model to handle images with parallax. They formulate the image warping as the thin plate spline model with a simple radial basis function.

Other spatially-varying warps \cite{chang2014shape,lin2015adaptive,li2017quasi,liu2019shape} focus on alleviating the distortions in the non-overlapping regions between input images, they often combine with APAP or other warping models to handle the parallax issues. The above warping methods can generate much better alignment results, however, they often fail to align the objects having large parallax and abrupt depth changes due to the smoothness constraint. 

Lee \emph{et al.} \cite{Lee_2020_CVPR} introduced the warping residual and estimated a discontinuous warp with ``holes'' to handle such issues. Lin \emph{et al.} \cite{lin2022image} proposed a method using a disparity map and multiple homographies to distinguish one background plane and multiple foreground objects and align them separately. The two methods both adopted iterative RANSAC to generate multiple homographies, while RANSAC can only provide \textit{sub-optimal} model to fit the inliers, and the effect of the outliers is hard to be eliminated and the random process may result in unstable homography models. Besides, the performance of Lee's method \cite{Lee_2020_CVPR} depends on the accuracy of the warping residual and the superpixel segmentation. The ``holes'' are produced via a hand-crafted selection strategy, which depends on the parameter setting. Lin's method \cite{lin2022image} cannot handle the case of feature match deficiency in the foreground objects. With the remarkable performance of SAM, our method can provide a more robust and accurate discontinuous warp.

%-------------------------------------------------------------------------
\subsection{Mesh-based Warps}

Another strategy to handle parallax is to represent the image as quad meshes and model the image warping as mesh deformation. Chen \emph{et al.} \cite{chen2016natural} proposed a mesh-based method with global similarity prior to address the distortions of unnatural rotation and scaling. They calculate the alignment term based on APAP-generated vertices correspondences instead of feature matches since the former is much more uniformly distributed. Li \emph{et. al} \cite{li2021image} and Zhang \emph{et. al} \cite{zhang2021natural} further developed the global similarity prior to handle the single image with multiple dominant regions.

%Zhang \emph{et. al} \cite{zhang2016multi}: They constrained the neighboring meshes to warp with similar homographies other than similarity transformations.
%Liao and Li \cite{liao2020Single}: They introduced both point and line feature matches to provide better alignment.
Zhang \emph{et. al} \cite{zhang2016multi} proposed to enforce a straightness constraint into the mesh deformation.  Liao and Li \cite{liao2020Single} proposed a single-perspective warping to mitigate the perspective distortions in the non-overlapping region. Jia \emph{et. al} \cite{jia2021Leveraging} further leveraged the line-point consistency to provide more accurate alignment and preserve both local and global lines. Chen \emph{et. al} \cite{chen2021Image} proposed an angle-consistent warping that integrates angle features of key points into homography estimation and mesh deformation. Du \emph{et. al} \cite{du2022Geometric} proposed a structure-preserving warping model to preserve the large-scale structures reflected by straight lines or curves.

The mesh-based warps focus on exploring different geometrical or content constraints to smoothly extrapolate the warping from overlapping region to non-overlapping region. For large parallax inputs, they easily fail.

\subsection{Other Warps}

To handle large parallax, a strategy called seam-driven is introduced which combines warping models with subsequent composition steps to generate visually pleasing results instead of geometrically accurate alignment. Gao \emph{et al.} \cite{gao2013seam} proposed to generate multiple homography candidates, and the homography bringing in the best seam-cutting result is chosen. Zhang \emph{et al.} \cite{zhang2014parallax} and Lin \emph{et al.} \cite{lin2016seagull} further developed the strategy by introducing mesh-based warping candidates which provide better alignment. The strategy aims to provide visually pleasing results, instead of the most geometrically accurate alignments. Besides, the warping candidates still need to align certain local regions well such that a plausible seam-cutting can be found. 

Some learning-based methods \cite{nie2020view,nie2021unsupervised} applied deep homography estimation \cite{detone2016deep} to stitch images. Nie \emph{et al.} \cite{nie2023parallax} further introduced an unsupervised stitching method which proposed a flexible warp to model the image alignment from global homography to local thin-plate spline motion. However, the flexible warp with smoothness constraint still fails to handle images with large parallax.

\section{Method}
\label{sec:method}

In this section, we propose our image stitching method via multi-homography warping, which mainly includes three steps: multi-homography fitting, multi-labeling for image contents, image warping and blending. The pipeline of our method is shown in Fig. \ref{fig:pipeline}.

\subsection{Multi-homography Fitting}

Given the target image and the reference image, we first segment the target image into numerous contents $\{C_k\}_{k=1}^M$ via SAM \cite{kirillov2023segment}. Then we detect the feature matches and remove the outliers via RANSAC \cite{fischler1981random}. Note that the feature matches cannot be fitted with a single homography due to the parallax. For this purpose, we employ the fundamental matrix to select the inliers based on the observation that a correct feature match $(\mathbf{p}, \mathbf{q})$ should satisfy:
\begin{equation*}
	\|\tilde{\mathbf{p}}^T F_* \tilde{\mathbf{q}}\|<\epsilon,
\end{equation*}
where $\tilde{\mathbf{p}}$ $(\tilde{\mathbf{q}})$ is the homogeneous coordinate of the feature point $\mathbf{p}$ $(\mathbf{q})$, and $F_*$ is obtained via solving the following equation using the direct linear transform (DLT) method:
\begin{equation}
	F_*=\arg\min_F \sum_{(\mathbf{p}, \mathbf{q})}\|\tilde{\mathbf{p}}^T F \tilde{\mathbf{q}}\|.
\end{equation}
%This results in 
The operation results in a set of feature matches $\{(\mathbf{p}_i, \mathbf{q}_i)\}_{i=1}^N$ between the input two images. Then we employ the multi-model fitting method \cite{isack2012energy} to partition the set of feature matches into several subsets, each of which fits a single homography. In particular, we aim to minimize the following energy function:
\begin{equation}
	E(\mathcal{H})=\sum_{i=1}^N \mathrm{D}(\mathbf{p}_i, H_{\mathbf{p}_i})
	+\lambda \sum_{(\mathbf{p}_i,\mathbf{p}_j)\in \mathcal{N}}\delta(H_{\mathbf{p}_i}\neq H_{\mathbf{p}_j})+\beta |\mathcal{H}|,
	\label{eq:1}
\end{equation}
where $\mathcal{H}=\{H_{\mathbf{p}_i}\ |\ \mathbf{p}_i\in P\}$ is the assignment of homography models to feature points $P=\{\mathbf{p}_i\}_{i=1}^N$. We next concretely explain the involved terms.

\noindent\textbf{The first term} $\mathrm{D}(\cdot,\cdot)$ measures the error of the feature point $\mathbf{p}_i$ fitted by the given homography $H_{\mathbf{p}_i}$, which is defined as the \textit{symmetric transfer error} (STE):
\begin{equation}
	\mathrm{D} (\mathbf{p}_i, H_{\mathbf{p}_i})=\|H_{\mathbf{p}_i}\mathbf{p}_i-\mathbf{q}_i\|+\|H^{-1}_{\mathbf{p}_i}\mathbf{q}_i-\mathbf{p}_i\|.	
\end{equation}
The homography model $H_{\mathbf{p}_i}$ is computed by minimizing the non-linear STE error of all the feature points fitted by $H_{\mathbf{p}_i}$ using the Levenberg-Marquardt method. The initial solution for the non-linear minimization is found using the DLT method.

It's worth noting that, the fundamental matrix-based RANSAC is ineffective in removing all the outliers with the obstruction of image noise. 
To properly separate the outliers from inliers in the multi-homography fitting step,
we define an extra ``outlier'' model $\emptyset$, where the model $\emptyset$ has constant fidelity measure $D(\mathbf{p}_i, \emptyset)=\gamma$ for all points $\mathbf{p}_i \in P$. 

\noindent\textbf{The second term} indeed measures the smoothness between the neighboring feature points fitted by different homography models. As the segmented contents of SAM provide abundant semantic information, we therefore construct a better neighborhood system of the feature points in the target image with such characteristics. Notice that any two feature points in the same content are parallax-free and can be fitted via the same homography, and the smoothness term in Eq.~\ref{eq:1} should be zero. To be specific, we define the neighborhood $\mathcal{N}$ as:
\begin{equation}
	\mathcal{N}=\{(\mathbf{p}_i, \mathbf{p}_j)\ |\ \mathbf{p}_i\in C_k^o,\ \mathbf{p}_j\in C_k^o,\ \overline{\mathbf{p}_i \mathbf{p}_j}\in \triangle_m\}_{k,m}
\end{equation}
where $C_k^o$ denotes the segmented content in the overlapping region $O$, $\triangle_m$ is the triangle generated by the Delaunay triangulation of the feature points. The above indicating function $\delta(\cdot)$ in Eq.~\ref{eq:1} is 1 if the condition inside the parenthesis is true and 0 otherwise.

\noindent\textbf{The third term} obviously controls the number of homography models.

% Furthermore, the fundamental matrix-based RANSAC is not effective in removing all the outliers with the obstruction of image noise. % and other noises. 
% To properly separate the outliers from inliers in the multi-homography fitting step,
% we define an extra ``outlier'' model $\emptyset$ as \cite{isack2012energy}, where the model $\emptyset$ has constant fidelity measure $D(\mathbf{p}_i, \emptyset)=\gamma$ for all points $\mathbf{p}_i \in P$. 

For the efficient optimization of the energy function in Eq.~\ref{eq:1}, the extension of the $\alpha$-expansion algorithm \cite{delong2012fast} is used.
For an accurate starting state, we first initialize the homography models $\mathcal{H}_0$ by iteratively performing the homography-based RANSAC until the number of the rest of the feature matches is smaller than a predefined threshold (50 in our experiment).

%Since each segmented content in SAM is semantical, we use it to construct a better neighborhood system of the feature points in the target image. Any two feature points in the same content are parallax-free and can be fitted via the same homography, thus the smoothness term should be zero. We define the neighborhood $\mathcal{N}$ as

\begin{figure}[t]
	\centering
	\subfloat[]{
		\includegraphics[height=0.095\textheight]{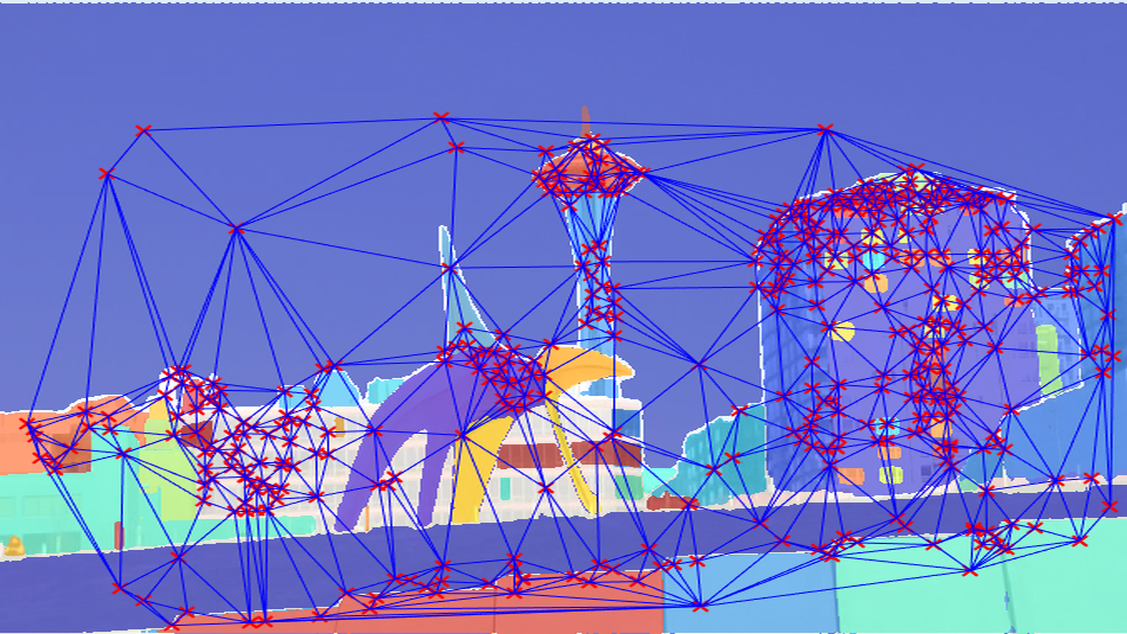}
		\includegraphics[height=0.095\textheight]{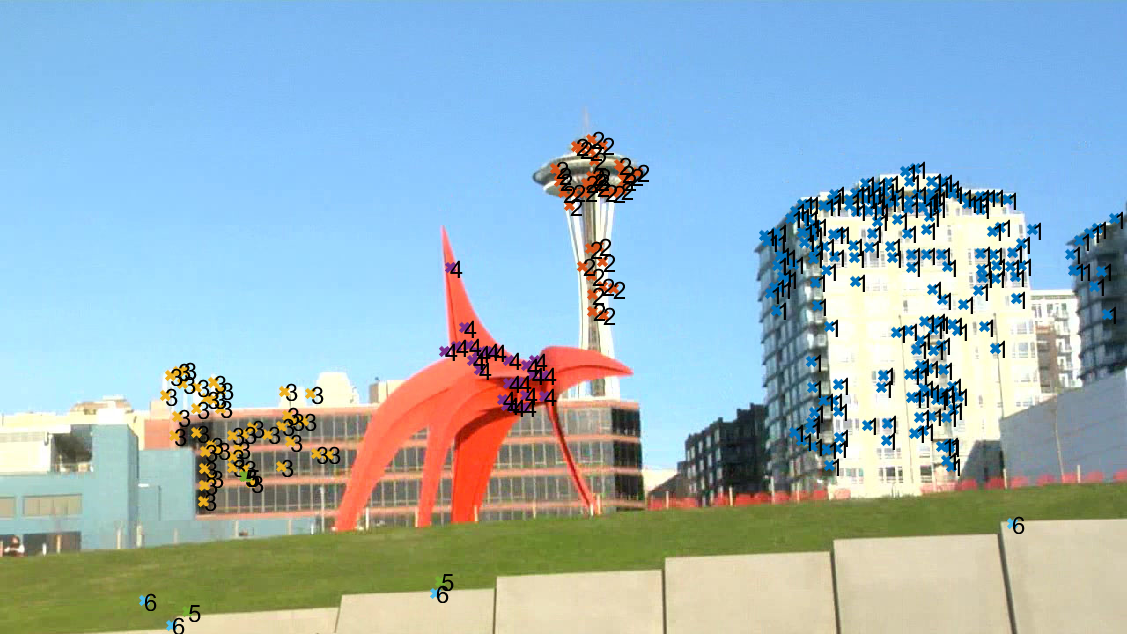}}\\
	\subfloat[]{	\includegraphics[height=0.095\textheight]{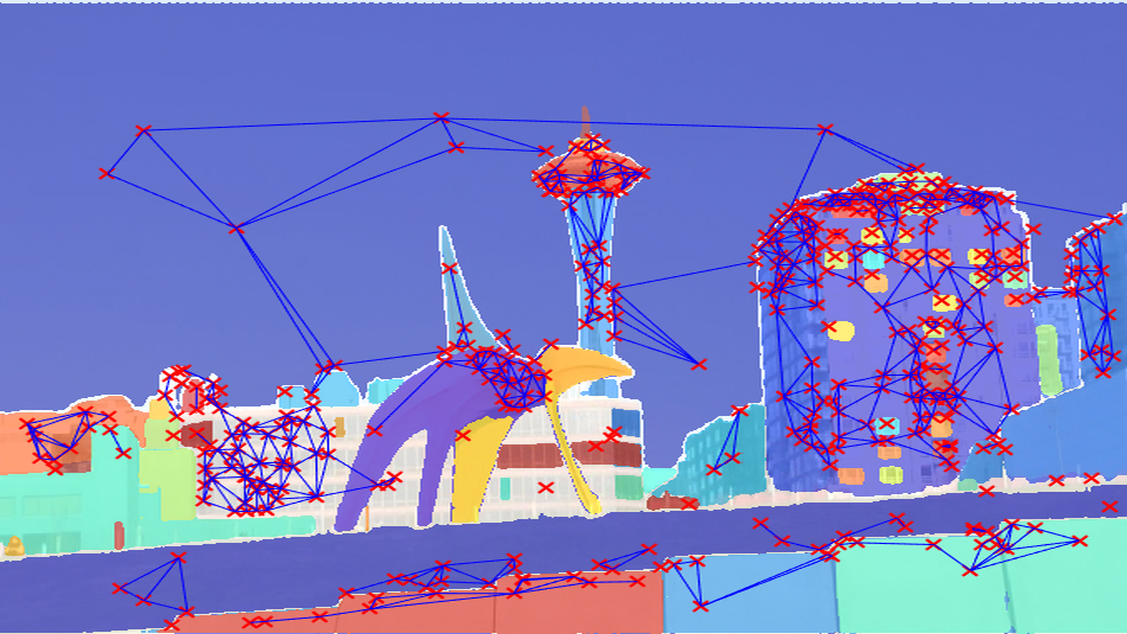}
		\includegraphics[height=0.095\textheight]{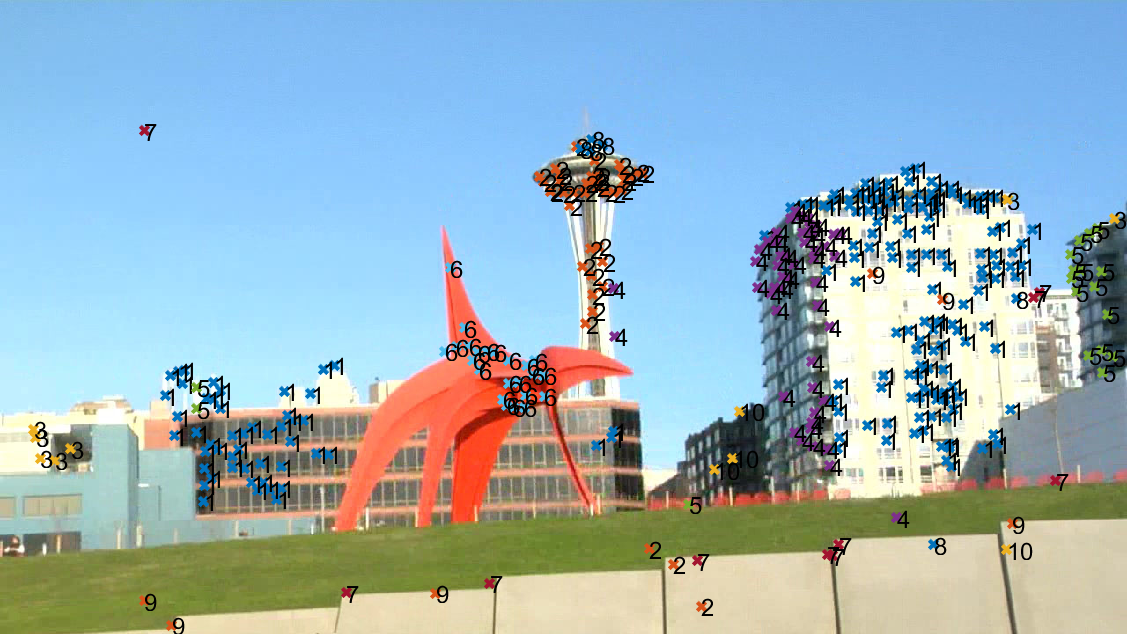}}\\
	\caption{Partition results of the feature points by our multi-homography fitting algorithm w/o and w/ SAM. (a) Neighborhood system defined only by Delaunay triangulation and the partition result. (b) Neighborhood system defined by Delaunay triangulation \& SAM, and the partition result. Introducing SAM to define the neighborhood system boosts the model performance.}
	\label{fig:neighborhood}
\end{figure}

\begin{figure*}[t]
	\centering
	\includegraphics[height=0.133\textheight]{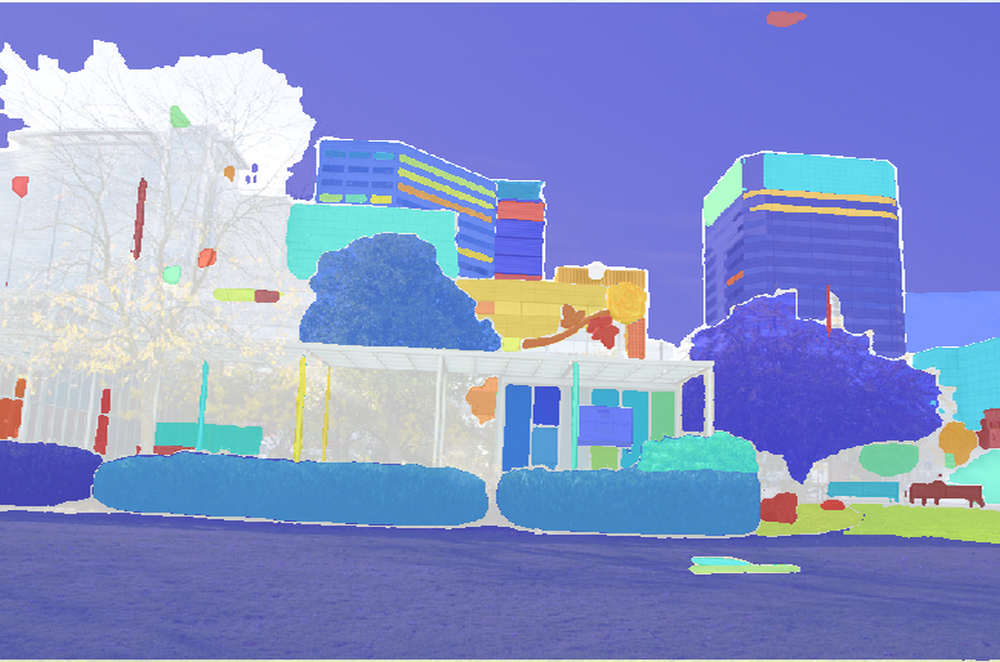}
	\includegraphics[height=0.133\textheight]{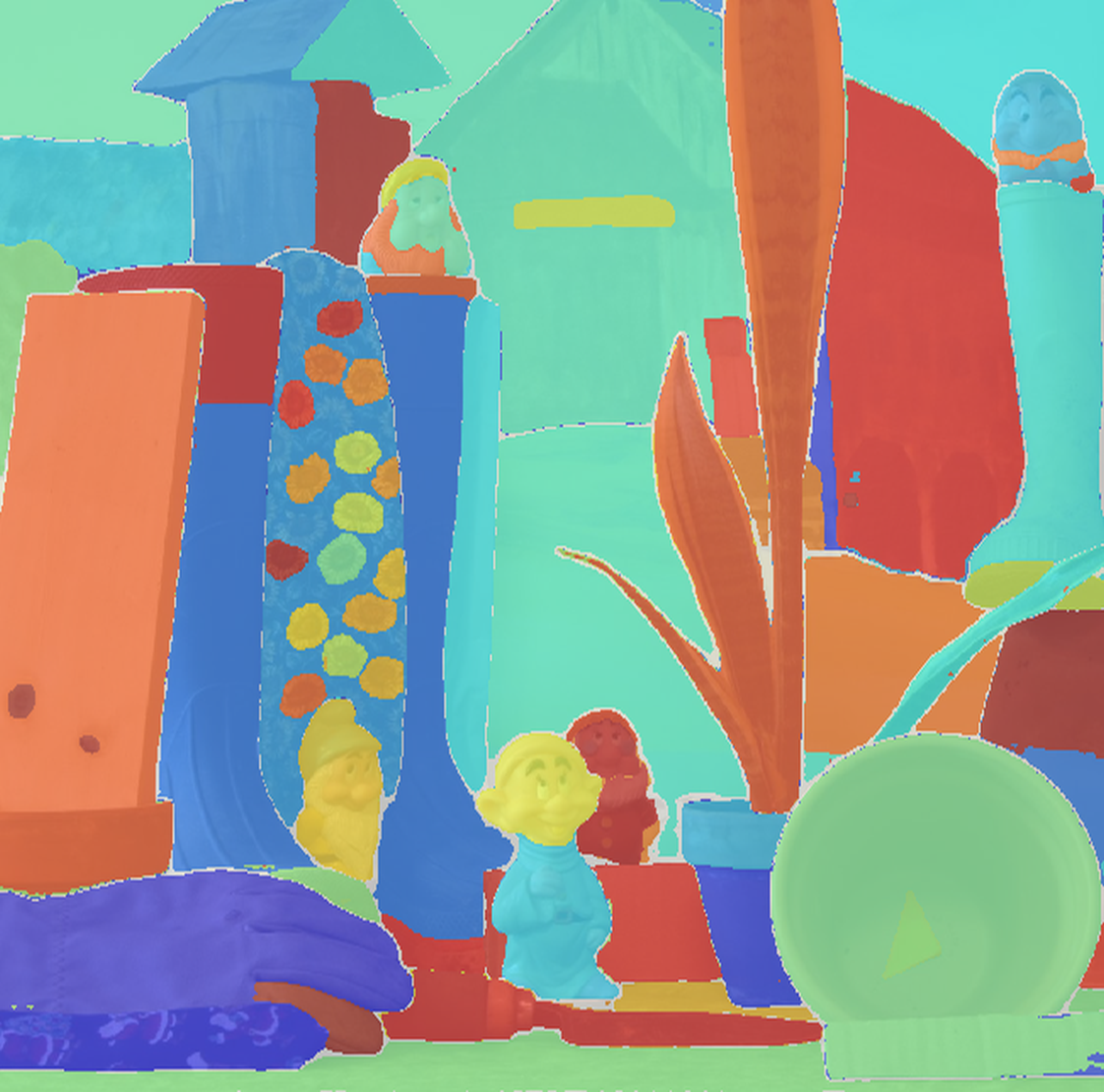}
	\includegraphics[height=0.133\textheight]{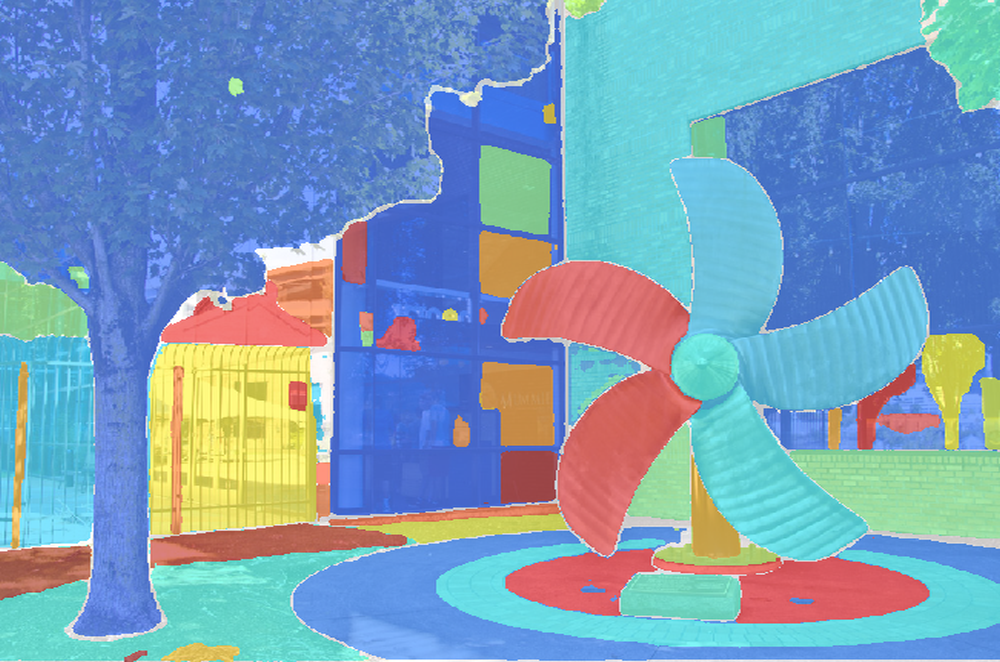}
	\includegraphics[height=0.133\textheight]{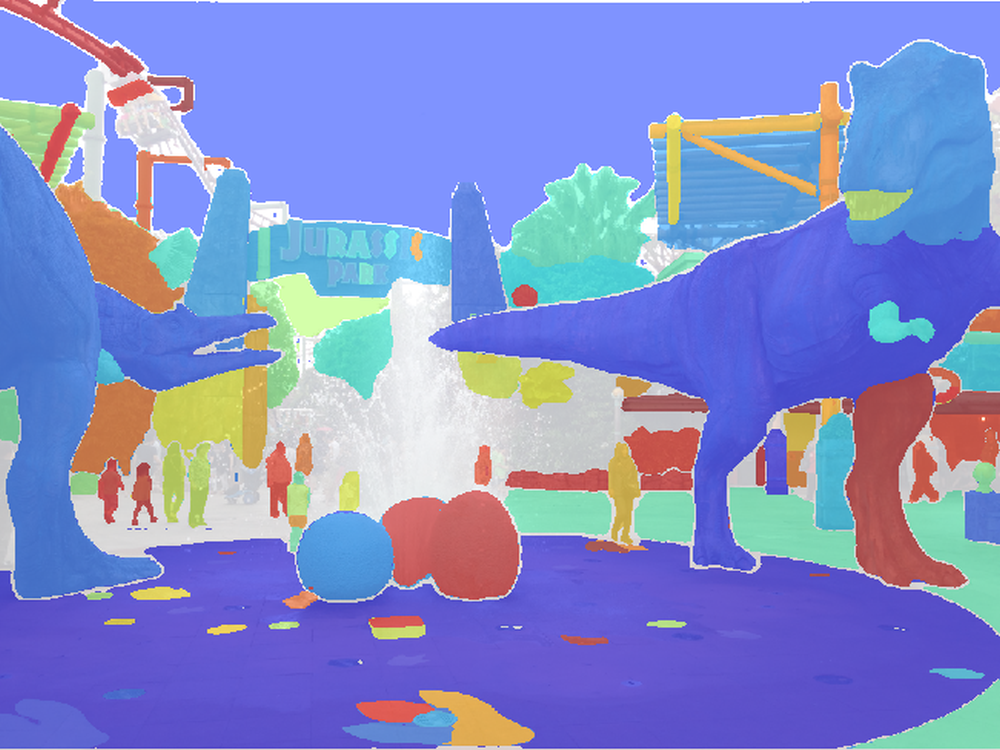}\\
	\includegraphics[height=0.133\textheight]{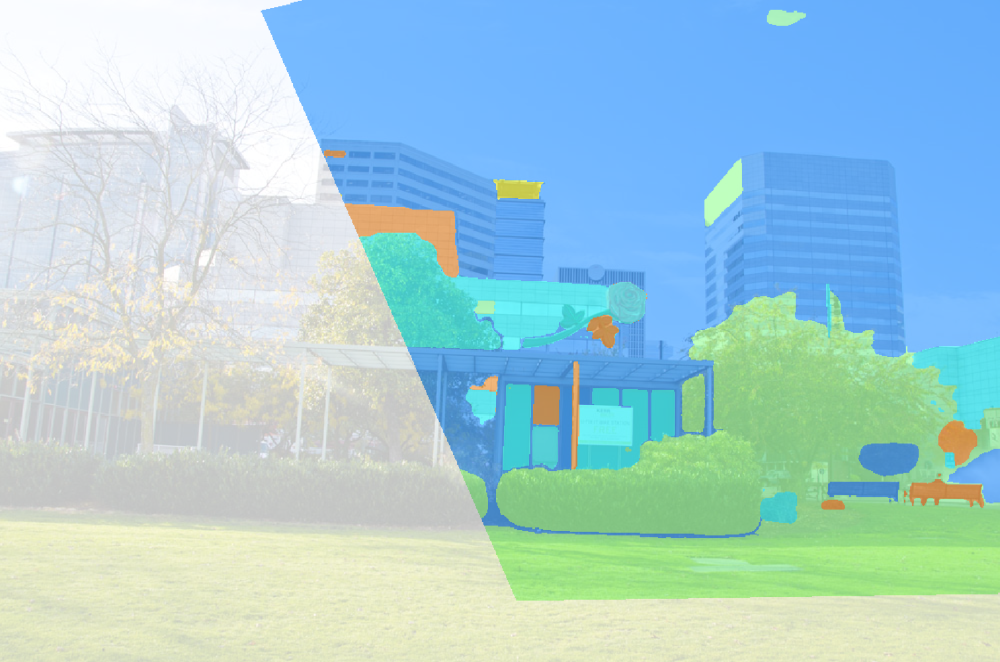}
	\includegraphics[height=0.133\textheight]{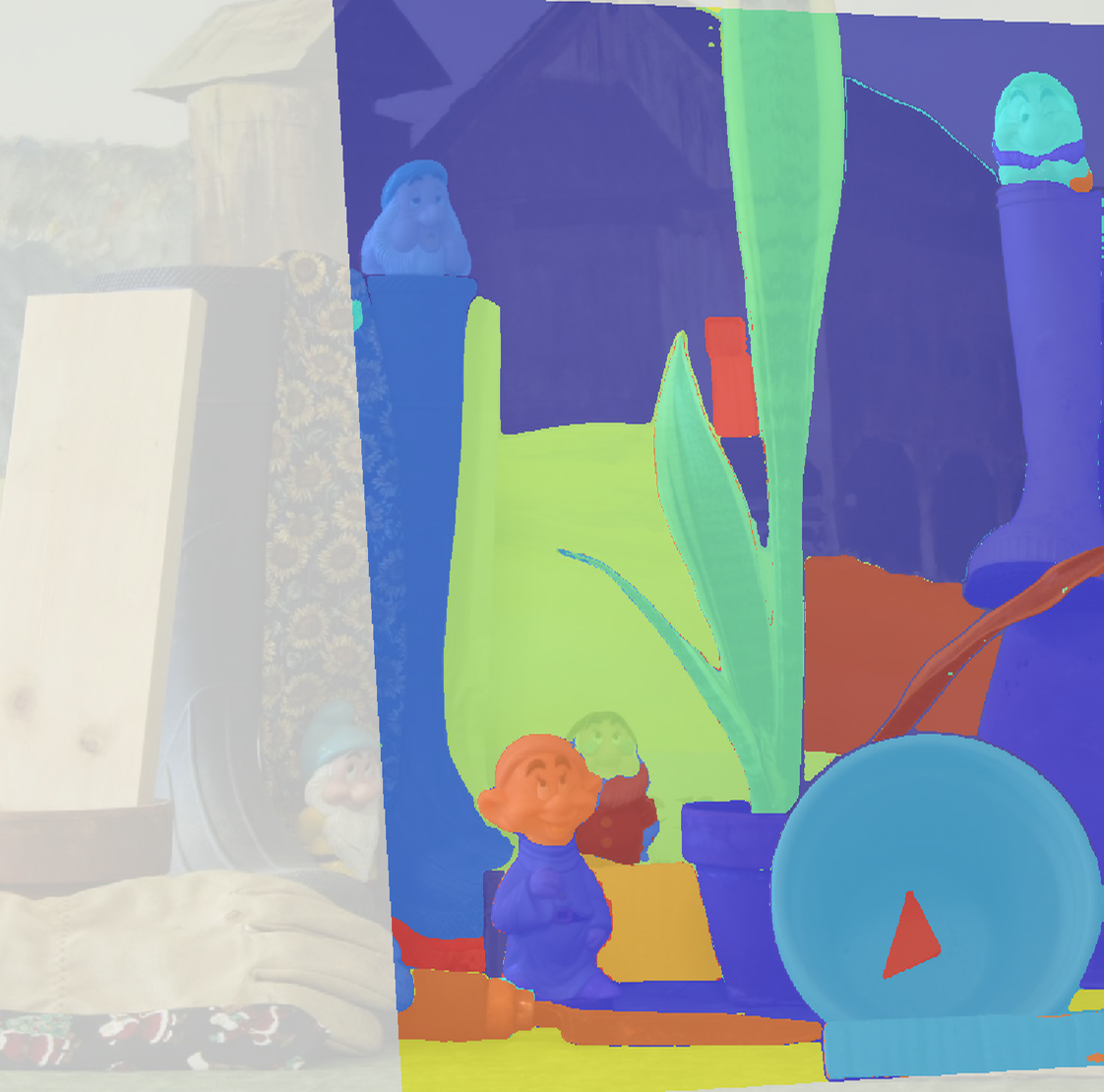}
	\includegraphics[height=0.133\textheight]{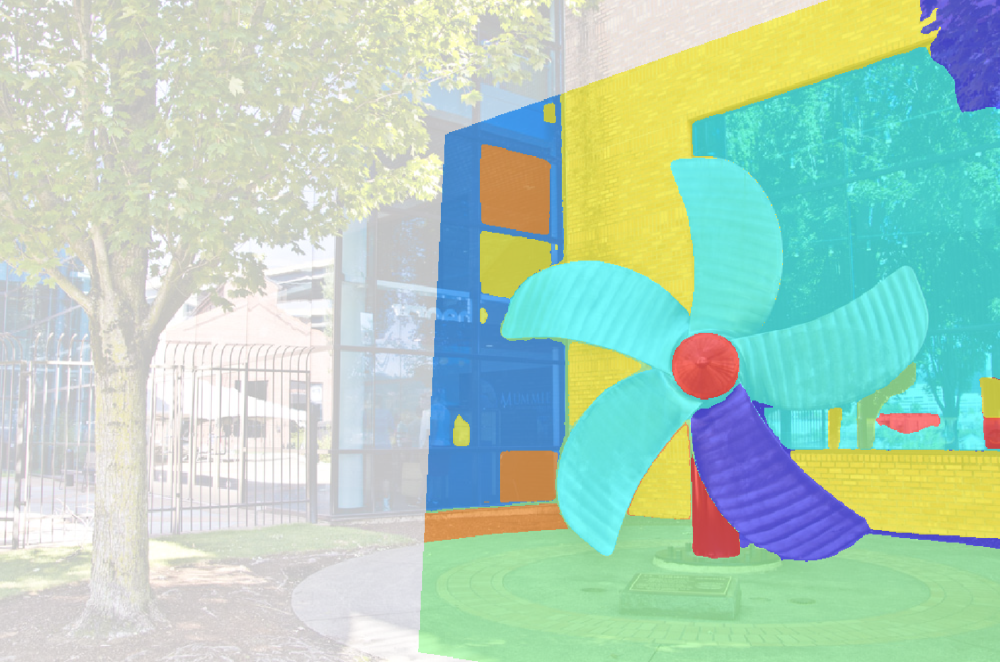}
	\includegraphics[height=0.133\textheight]{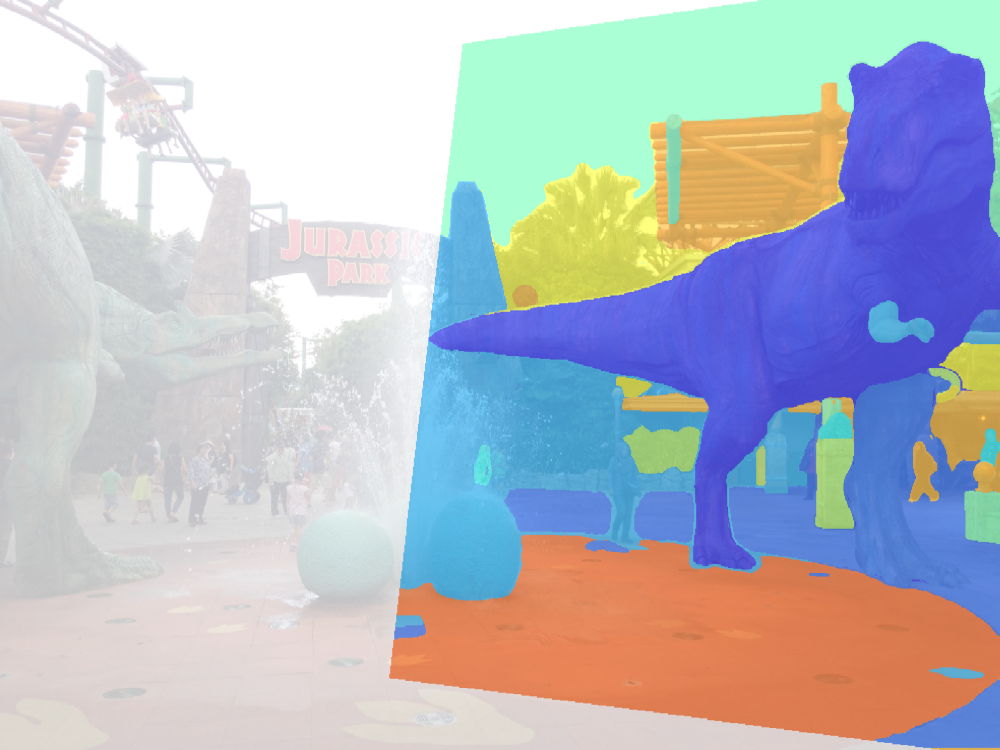}\\
	\includegraphics[height=0.133\textheight]{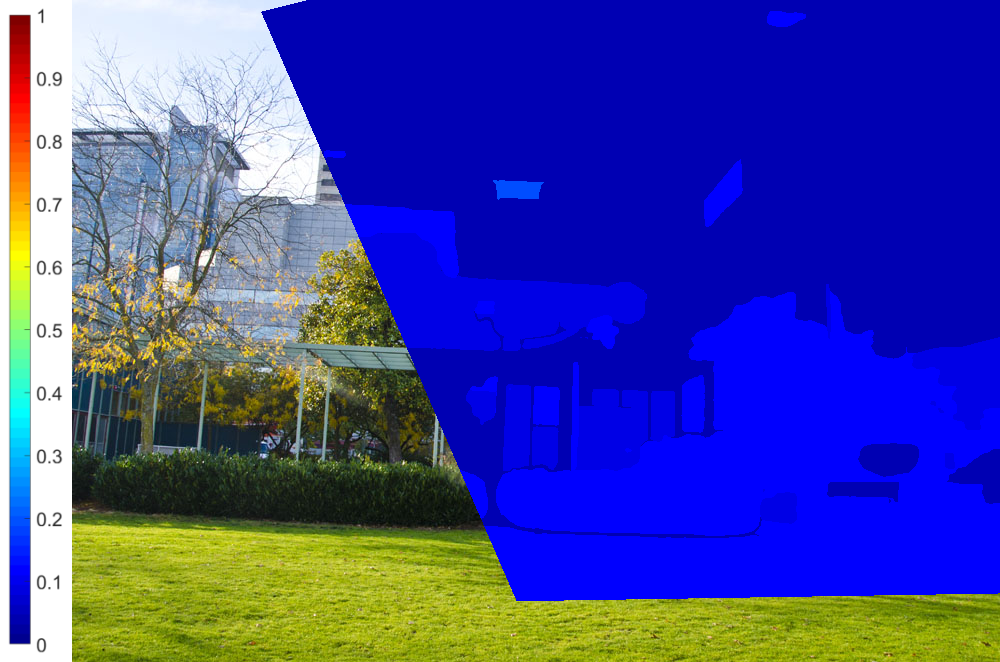}
	\includegraphics[height=0.133\textheight]{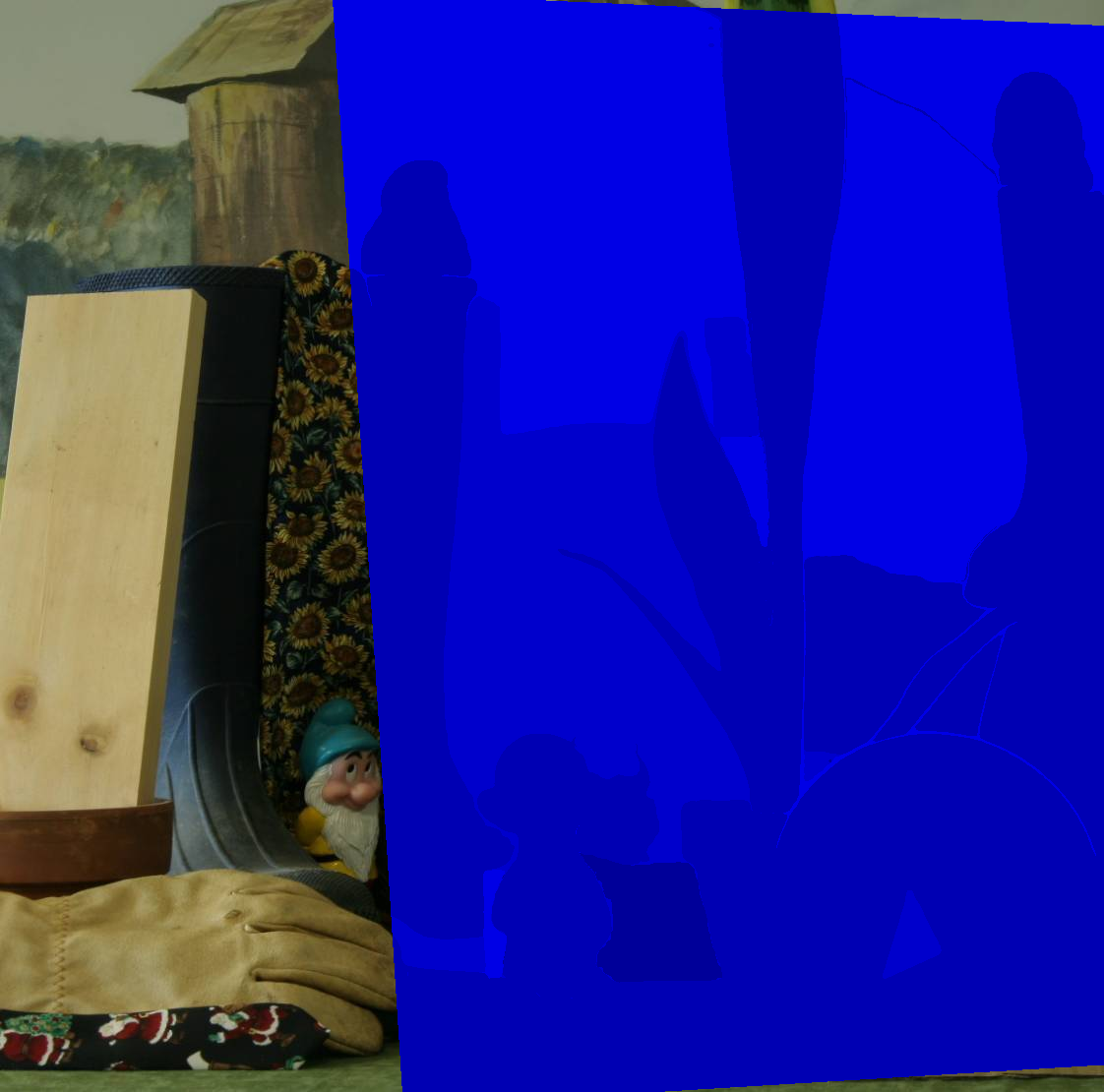}
	\includegraphics[height=0.133\textheight]{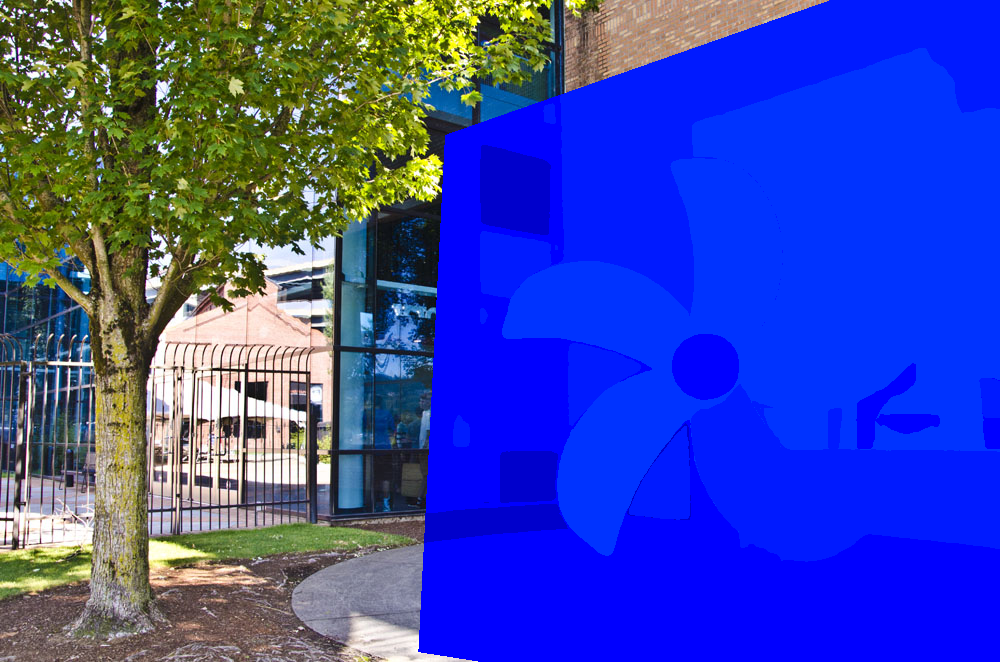}
	\includegraphics[height=0.133\textheight]{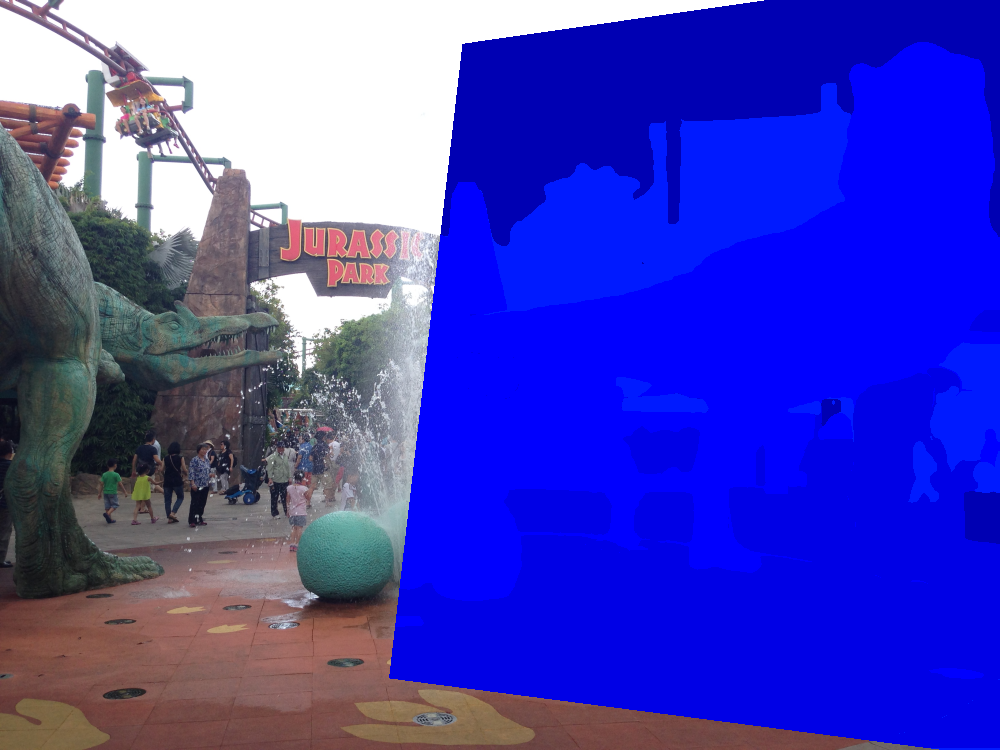}\\
	\caption{Labeling results on different image pairs. \textbf{Top}: SAM results. \textbf{Middle}: labeling results for overlapping regions. Each color corresponds to a label and fits a homography. \textbf{Bottom}: average photometric errors for each labeling content, where errors are shown as a hot map.}
	\label{fig:labeling}
\end{figure*}

For ease of understanding, we here visualize one multi-homography fitting result on the test case \cite{zhang2014parallax} in Fig. \ref{fig:pipeline}. Ten subsets of feature matches are partitioned, each fitting a homography. Among them, the first homography successfully aligns the background buildings. For the foreground scene, the second and the sixth fitted homographies align the tower and the red statue, respectively. To verify the effectiveness of SAM-based contents, we compare the fitting results calculated by defining the neighborhood system w/o and w/ SAM-based contents. 
%We also compare with the fitting result calculated by defining the neighborhood system only by Delaunay triangulation, as shown in Fig. \ref{fig:neighborhood}. 
As shown in Fig.~\ref{fig:neighborhood}, the neighborhood system involving SAM can produce a more accurate partition of the feature points. Sec.~\ref{sec:5-3} will provide a detailed discussion to prove its effectiveness.

\subsection{Multi-labeling for Image Contents}

Then, we use the obtained multiple homographies to label (align) the segmented contents of the target image. 

\noindent\textbf{Labeling for the overlapping region.} We first calculate a global homography warp $H_g$ using all the feature matches to construct the overlapping region. Then, for each content $C^o_k$ in the overlapping region, we label it using the homography that aligns the reference image with the lowest photometric error, and the photometric error of the content $C^o_k$ aligned by $H_i$ is calculated as:
\begin{equation}
	e(C_k^o, H_i)=\frac{1}{|C^o_k|}\sum_{\mathbf{p}\in C^o_k}\|I_r(H_i(\mathbf{p}))-I_t(\mathbf{p})\|,\ H_i\in \mathcal{H}.
	\label{eq:2}
\end{equation}
where $H_i$ maps the pixel coordinate $\mathbf{p}$ in the target image to the coordinate $H_i(\mathbf{p})$ in the reference image.

%%-- Labeling for non-overlapping region.
\noindent\textbf{Labeling for the non-overlapping region.}
To smoothly extrapolate the warp
from the overlapping region to the non-overlapping region,
we linearize the multiple homographies in the overlapping region and utilize them to label the image contents in the non-overlapping region, which is inspired by~\cite{lin2015adaptive}. Moreover, we use a similarity transformation to reduce the perspective distortion in the non-overlapping region. 
%we adopt the same strategy as \cite{lin2015adaptive} by linearizing the homography transformation and using it to label each content $S^n_k$ in the non-overlapping region. 
In particular, we use each subset of feature matches to calculate an individual similarity transformation and choose the one with the smallest rotation angle. Then, we  uniformly sample $R_1$ anchor points $\{\mathbf{a}_i\}_{i=1}^{R_1}$ in the boundary of the overlapping region and $R_2$ anchor points $\{\mathbf{b}_j\}_{j=1}^{R_2}$ in the outermost boundary of the non-overlapping region. Then we partition the non-overlapping region into a grid of cells and take the center of each cell as $\mathbf{c}_k$. The linearized homography for the cell centering at $\mathbf{c}_k$ is calculated as
\begin{align}
	H_{\mathbf{c}_k} & = \sum_{i=1}^{R_1} \alpha_i \left(H_{\mathbf{a}_i} + J_H(\mathbf{a}_i)(\mathbf{c}_k-\mathbf{a}_i)\right)\nonumber\\
	& + \sum_{j=1}^{R_2} \alpha_j \left(\mathbf{S} + J_\mathbf{S}(\mathbf{b}_j)(\mathbf{c}_k-\mathbf{b}_j)\right)
	\label{eq:nonoverlapping}
\end{align}
where $J_H(\mathbf{a}_i)$ is the Jacobian of the homography $H_{\mathbf{a}_i}$ at the point $\mathbf{a}_i$, $J_\mathbf{S}(\mathbf{b}_j)$ is the Jacobian of the optimal similarity transformation $\mathbf{S}$ at the point $\mathbf{b}_j$. $\alpha_i$ ($\alpha_j$) is a function of $\mathbf{c}_k$ and $\mathbf{a}_i$ ($\mathbf{b}_j$), defined by Student's t-weighting:
\begin{equation}
	\alpha_i = \left(1+\frac{\|\mathbf{c}_k-\mathbf{a}_i\|^2}{\nu}\right)^{\frac{-(\nu+1)}{2}}
\end{equation}	
\begin{equation}	
	\alpha_j = \left(1+\frac{\|\mathbf{c}_k-\mathbf{b}_j\|^2}{\nu}\right)^{\frac{-(\nu+1)}{2}}
\end{equation}
where $\nu$ controls the decaying rate of $\alpha_i$ ($\alpha_j$). Specifically, $\alpha_i$ assigns a higher weight to the cell in the neighborhood of the anchor point $\mathbf{a}_i$, making the linearized homography $H_{\mathbf{c}_k}$ similar to $H_{\mathbf{a}_i}$ and gradually transform to $\mathbf{S}$. Thus, the multiple homographies can be smoothly extrapolated to the non-overlapping region. We calculate the linearized homographies for dense cells instead of the segmented contents in the non-overlapping region because the former can generate smooth and matched boundaries between the neighboring cells, while the segmented contents don't.

\begin{figure}[t]
	\centering
	\subfloat[]{
		\includegraphics[height=0.104\textheight]{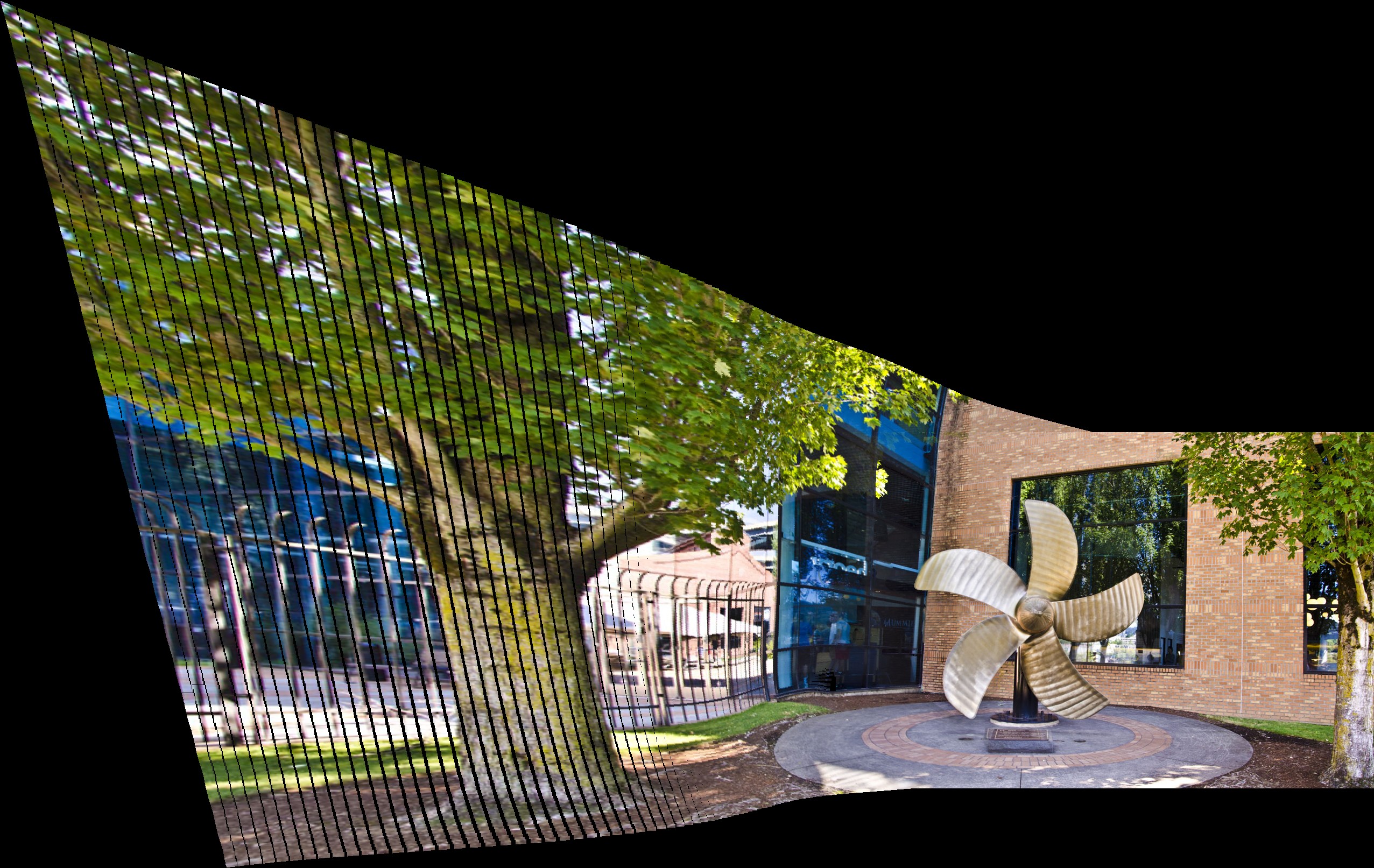}}
	\subfloat[]{
		\includegraphics[height=0.104\textheight]{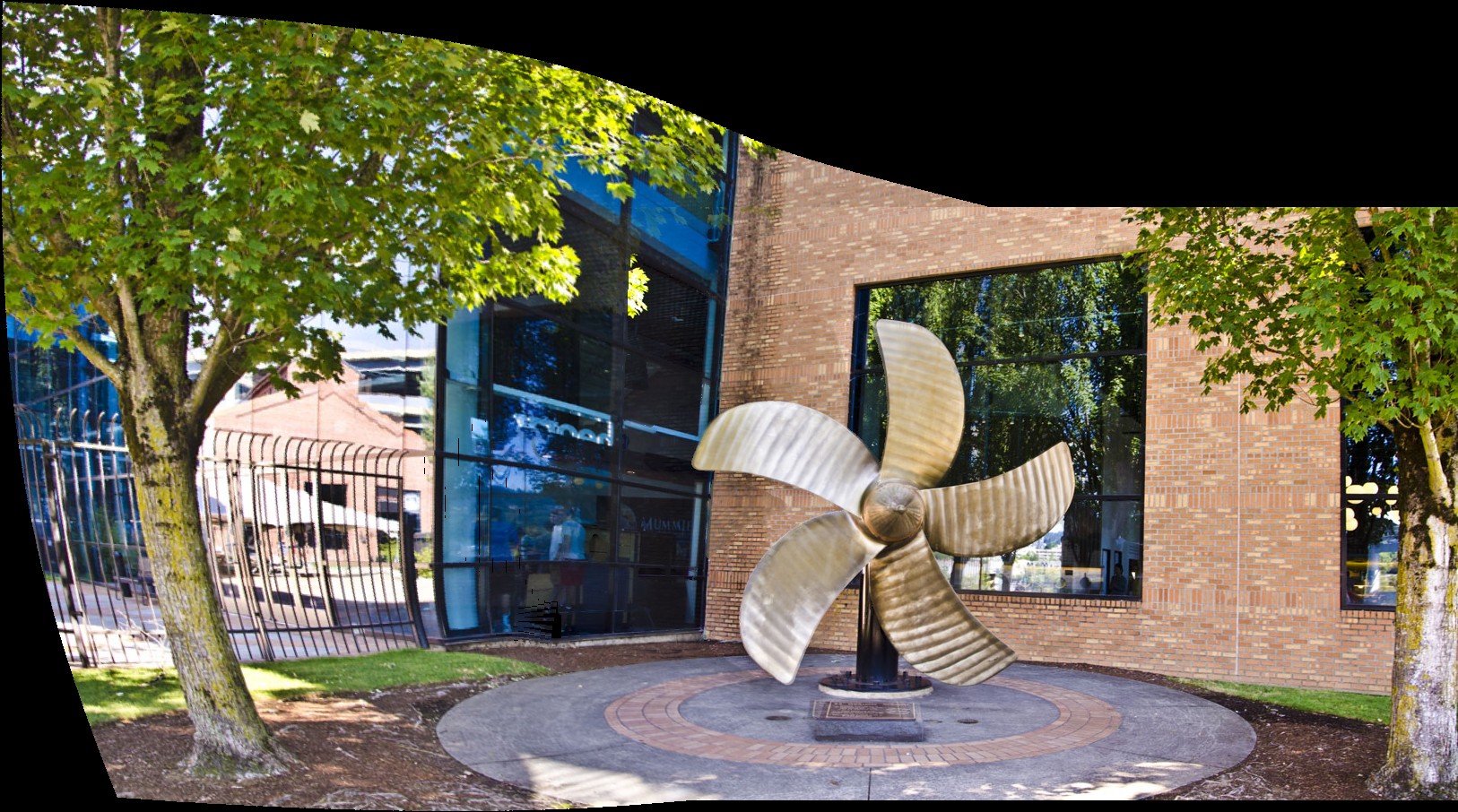}}\\
	\subfloat[]{
		\includegraphics[height=0.098\textheight]{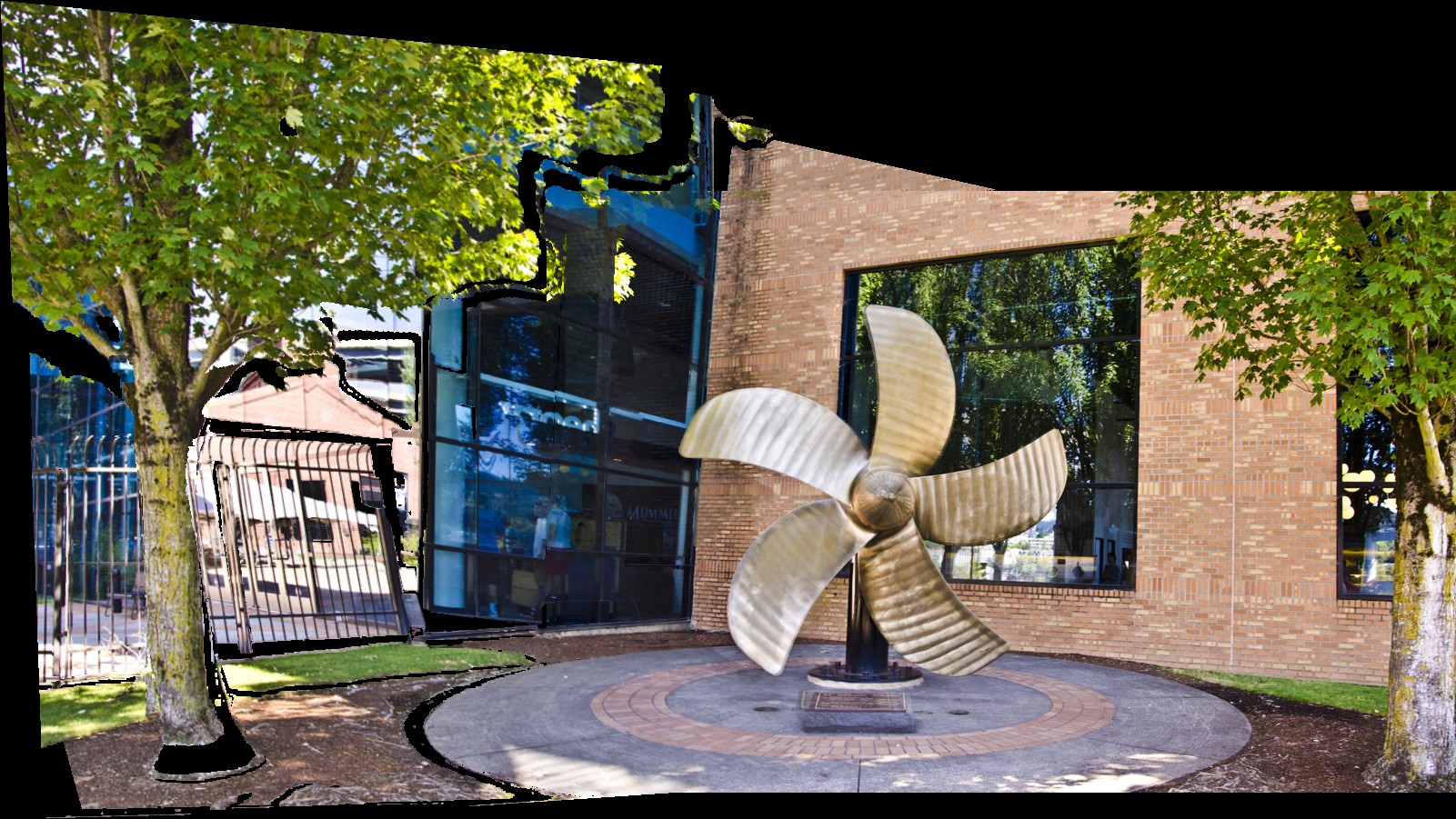}}
	\subfloat[]{
		\includegraphics[height=0.098\textheight]{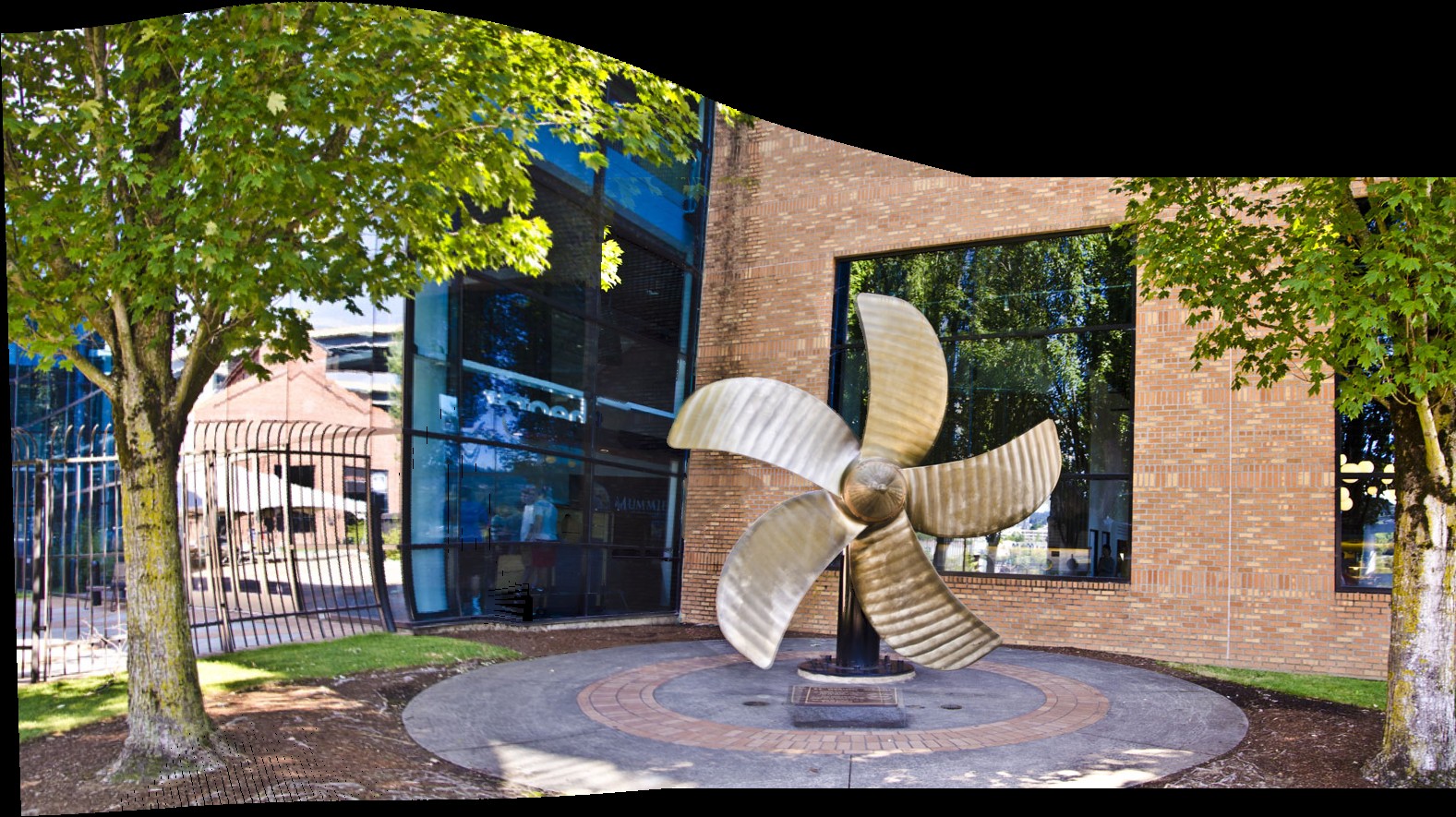}}\\
	\caption{Different Labeling strategies for the non-overlapping region. (a) Labeling strategy which replaces $\mathbf{S}$ with global homography $H_g$. (b) Labeling strategy only using the first term of Eq. \ref{eq:nonoverlapping}. (c) Labeling strategy which replaces dense cells with segmented contents by SAM. (d) Our final labeling strategy.}
	\label{fig:nonoverlaping}
\end{figure}

%for any point $\mathbf{r}$ in the neighborhood of the anchor point $\mathbf{a}$, its linearized homography can be considered as the first-order Taylor series,
%\begin{equation}
%	H_\mathbf{r} = H_\mathbf{a} + J_H(\mathbf{a})(\mathbf{r}-\mathbf{a})
%\end{equation}
%where $J_H(\mathbf{a})$ is the Jacobian of the homography $H$ at the point $\mathbf{a}$. 

Recalling the visualization result in Fig.~\ref{fig:pipeline}, we further demonstrate the multi-labeling result, where each color represents a label and corresponds to a homography. More labeling results are also exhibited in Fig.~\ref{fig:labeling}, which contains several SAM results, labeling results in the overlapping region, and the colored photometric errors calculated by Eq. (\ref{eq:2}). The labeling homography for each content in the overlapping region provides very small errors. We also show the stitching results using different labeling strategies for the non-overlapping region in Fig. \ref{fig:nonoverlaping}. Including similarity transformation into homography linearization for dense cells reduces the perspective distortions and achieves the best stitching quality.

\subsection{Image Warping and Blending}

The multi-homography fitting and multi-labeling provide multiple homographies for corresponding contents $C_k^o$ in the overlapping region and grid cells $\mathbf{c}_k$ in the non-overlapping region. Finally, we warp the target image to align with the reference image and composite them via linear blending. Here we use both forward and backward mapping to generate the warped target and reference images. 

%\begin{figure}[t]
%	\centering
%	\includegraphics[width=0.4\textwidth]{texture_mapping.pdf}
%	\caption{A simple illustration of our image warping process. $H_1$ and $H_2$ are two fitted homographies for the image contents in the overlapping region, $L(H_1, H_2)$ and $L(H_2, H_1)$ are the linearized homographies with different weight combinations for the non-overlapping region.}
%	\label{fig:mapping}
%\end{figure}

The forward mapping is capable of accurately mapping each labeling object to the corresponding object in the reference image. While there exist float coordinate issues in the mapping process. We address it by combining backward mapping on each labeling region. Specifically, we use forward mapping to calculate the canvas and backward texture mapping for each labeling region to render the image.
%The illustration in Fig.~\ref{fig:mapping} exhibits how we generate the warped target image, where we use forward mapping to calculate the canvas and backward texture mapping for each labeling region to render the image.

\begin{figure}[t]
	\centering
	\subfloat[]{
		\includegraphics[height=0.117\textheight]{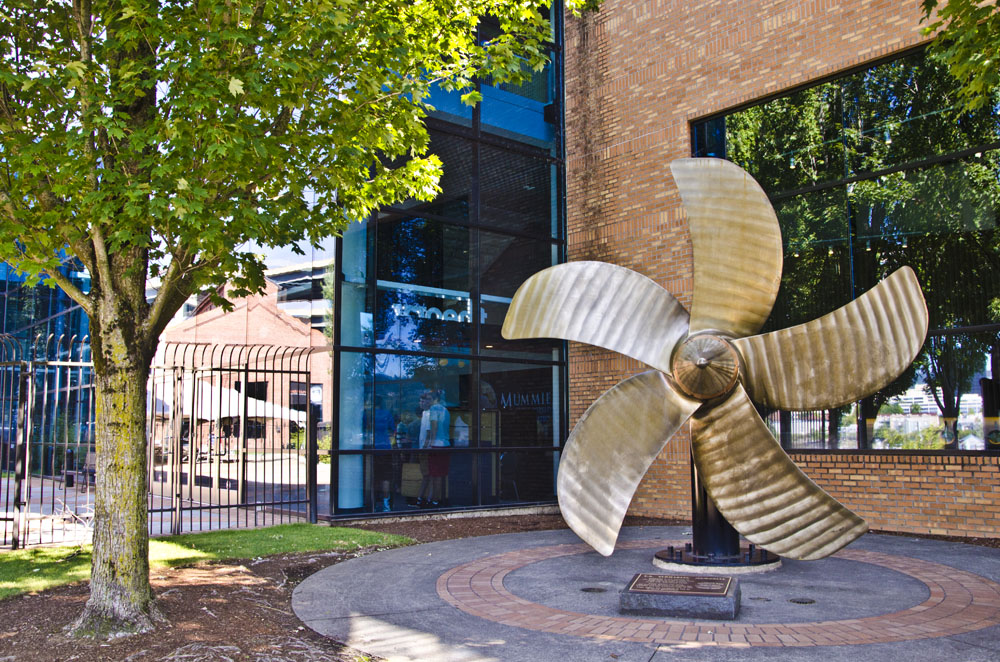}
		\includegraphics[height=0.117\textheight]{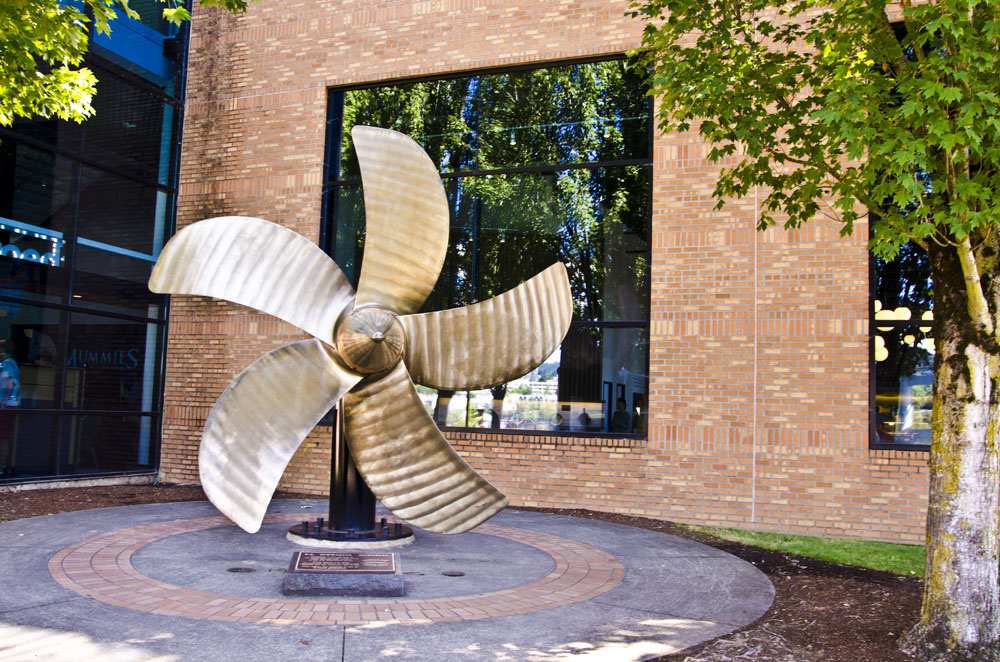}}\\
	\subfloat[]{
		\includegraphics[height=0.1\textheight]{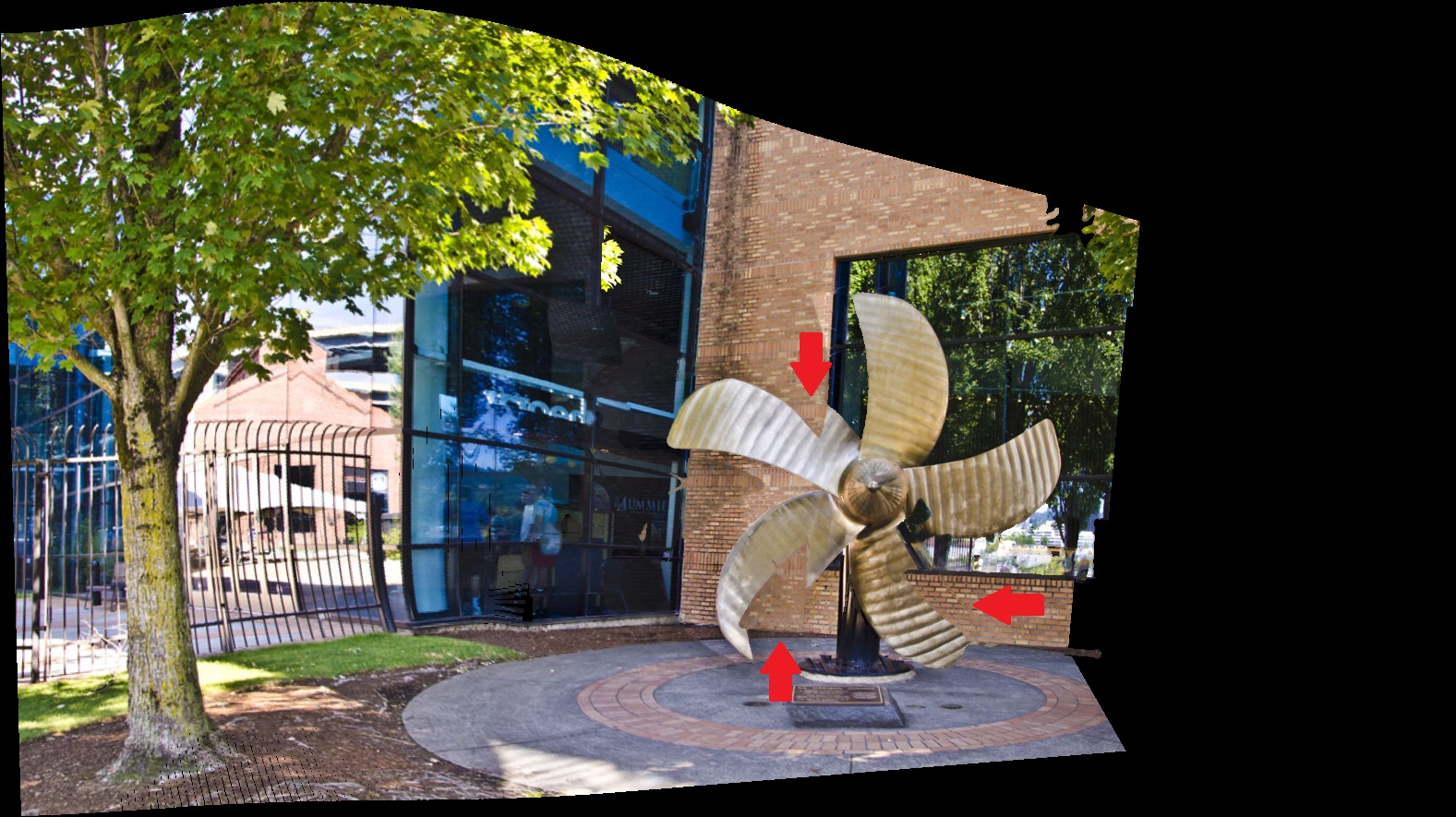}}
	\subfloat[]{	
		\includegraphics[height=0.1\textheight]{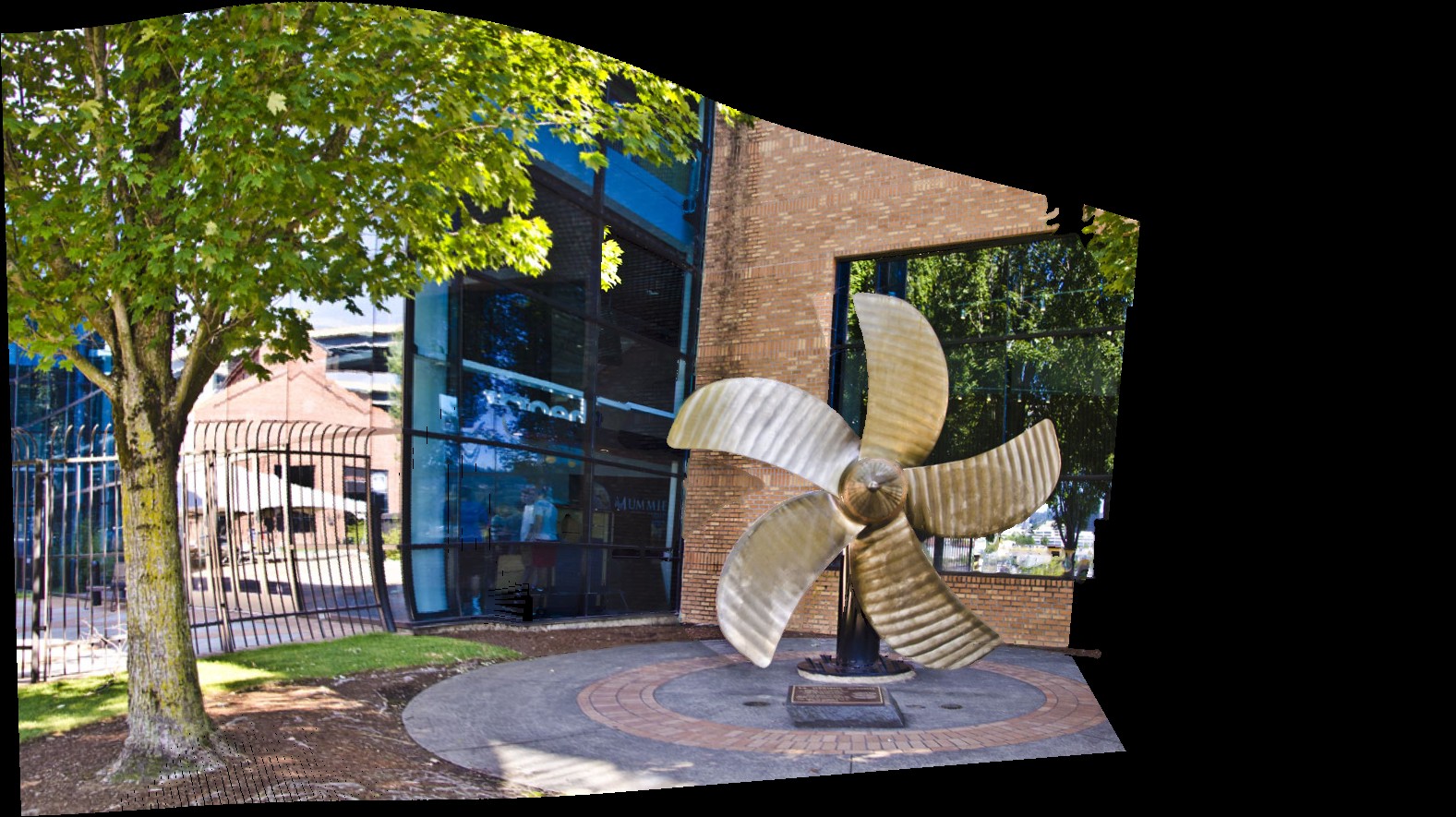}}\\
	\caption{The occlusion issue in the texture mapping. (a) Input images. (b) Warped image without error-buffer. (c) Warped image with error-buffer. Red arrows indicate the occlusion issue in warped images.}
	\label{fig:overlapped}
\end{figure}

Besides, due to the parallax between foreground objects and scenes, the input images may have different occlusion relations, which results in the ``n-to-1'' forward mapping issue. We further deal with it by introducing a ``\textbf{Error-buffer}'' algorithm. Specifically, for $n$ regions in the target image that are mapped to the same region in the reference image, we compute the $n$ average photometric errors and warp the region with the smallest error to align with the reference image. Fig.~\ref{fig:overlapped} shows an example of the occlusion issue. With the introduction of the error-buffer algorithm, the warped target image is well-rendered with the same occlusion relation as the reference image.

\section{Experiments}
\label{sec:exp}

\noindent\textbf{Settings.} We conduct comparative experiments of the proposed method on a variety of existing available datasets with large parallax, including 35 image pairs from Parallax \cite{zhang2014parallax}, 24 image pairs from SEAGULL \cite{lin2016seagull} and 14 image pairs from MR \cite{herrmann2018robust}. For complete comparisons, the state-of-the-art methods, including single homography (Baseline), APAP \cite{zaragoza2014projective}, SPHP \cite{chang2014shape}, ANAP \cite{lin2015adaptive}, GSP \cite{chen2016natural}, REW \cite{li2018parallax}, SPW \cite{liao2020Single}, TFA \cite{li2019local}, LPC \cite{jia2021Leveraging} and UDIS++ \cite{nie2023parallax} are involved. The model parameters of existing methods are set as the original papers. To highlight the alignment performance, all the warping results are composited using linear blending. In the following experiments, feature points are detected and matched using SIFT \cite{lowe2004distinctive}. For our model parameters, we set $\lambda=20, \beta=10, \gamma=200$ for multi-homography fitting, and $\nu$ is set to 5 in Student's t-weighting. The proposed method typically takes from 5 to 10 seconds\footnote{The inference time includes SAM, SIFT feature detection and matching, multi-homography warping and final linear blending.} with a 3.6GHz CPU and 16GB RAM to stitch two images with a resolution of 1000$\times$750.

\subsection{Quantitative Comparison}

To evaluate the alignment performance of these warping methods, we introduce three 
Image Quality Assessment (IQA) metrics, PSNR, SSIM \cite{wang2004image}, and LPIPS \cite{zhang2018unreasonable}, to compare the alignment quality. The three metrics are calculated based on the overlapping regions of warped images. The average metrics on the three datasets are shown in Table.~\ref{tab:quality}. Note that in a few image pairs, some methods fail to produce any meaningful results, thus we don't include these cases in their average metric calculation, while additionally counting them as the number of Failure Cases (\# FC). Our method consistently achieves the best alignment performance by a large margin on the three datasets according to the involved metrics.

\begin{table*}
	%\scriptsize
	\begin{center}
		\caption{Quantitative comparisons between SOTA warping methods on different challenge datasets. The best is marked in bold and the second best is in underline.} \label{tab:quality}%
		\begin{tabular}{l|ccc|ccc|ccc|c}
			\hline
			& \multicolumn{3}{c|}{Parallax \cite{zhang2014parallax}} & \multicolumn{3}{c|}{SEAGULL \cite{lin2016seagull}} & \multicolumn{3}{c|}{MR \cite{herrmann2018robust}}\\
			Method   & PSNR $\uparrow$ & SSIM $\uparrow$  & LPIPS $\downarrow$ & PSNR $\uparrow$  & SSIM $\uparrow$  & LPIPS $\downarrow$ & PSNR $\uparrow$  & SSIM $\uparrow$  & LPIPS $\downarrow$ & \# FC\\
			\hline
			Baseline & 15.37 & 0.614 & 0.297 & 15.93 & 0.592 & 0.313 & 14.57 & 0.565 & 0.325 & 3\\
			APAP \cite{zaragoza2014projective} & 16.84 & 0.659 & 0.237 & 16.63 & 0.613 & 0.282 & 15.89 & 0.608 & 0.282 & 0 \\
			SPHP \cite{chang2014shape} & 15.77 & 0.662 & 0.317 & 15.90 & 0.640 & 0.339 & 15.02 & \underline{0.664} & 0.364 & 1 \\
			ANAP \cite{lin2015adaptive} & 16.87 & 0.666 & 0.227 & 17.00 & 0.640 & 0.247 & \underline{16.11} & 0.619 & \underline{0.269} & 1 \\
			GSP \cite{chen2016natural} & 17.30 & 0.698 & \underline{0.215} & \underline{17.34} & \underline{0.676} & \underline{0.228} & 15.28 & 0.621 & 0.274 & 0 \\
			REW \cite{li2018parallax} & \underline{17.37} & \underline{0.700} & 0.233 & 16.78 & 0.650 & 0.290 & 14.78 & 0.587 & 0.339 & 0 \\
			SPW \cite{liao2020Single} & 16.33 & 0.642 & 0.250 & 16.49 & 0.602 & 0.285 & 15.49 & 0.574 & 0.291 & 0 \\
			TFA \cite{li2019local} & 16.00 & 0.637 & 0.319 & 16.63 & 0.650 & 0.293 & 14.45 & 0.567 & 0.366 & 1 \\
			LPC \cite{jia2021Leveraging} & 16.33 & 0.634 & 0.256 & 16.03 & 0.588 & 0.299 & 14.29 & 0.514 & 0.342 & 0 \\
			UDIS++ \cite{nie2023parallax} & 15.64 & 0.606 & 0.269 & 16.09 & 0.577 & 0.294 & 15.02 & 0.542 & 0.311 & 1 \\
			Ours  & \textbf{19.19} & \textbf{0.752} & \textbf{0.192} & \textbf{18.69} & \textbf{0.713} & \textbf{0.226} & \textbf{18.55} & \textbf{0.736} & \textbf{0.197} & 0 \\
			\hline
		\end{tabular}%
	\end{center}
\end{table*}

\subsection{Panorama Comparison}

\begin{figure*}[t]
	\centering
	\includegraphics[width=0.98\textwidth]{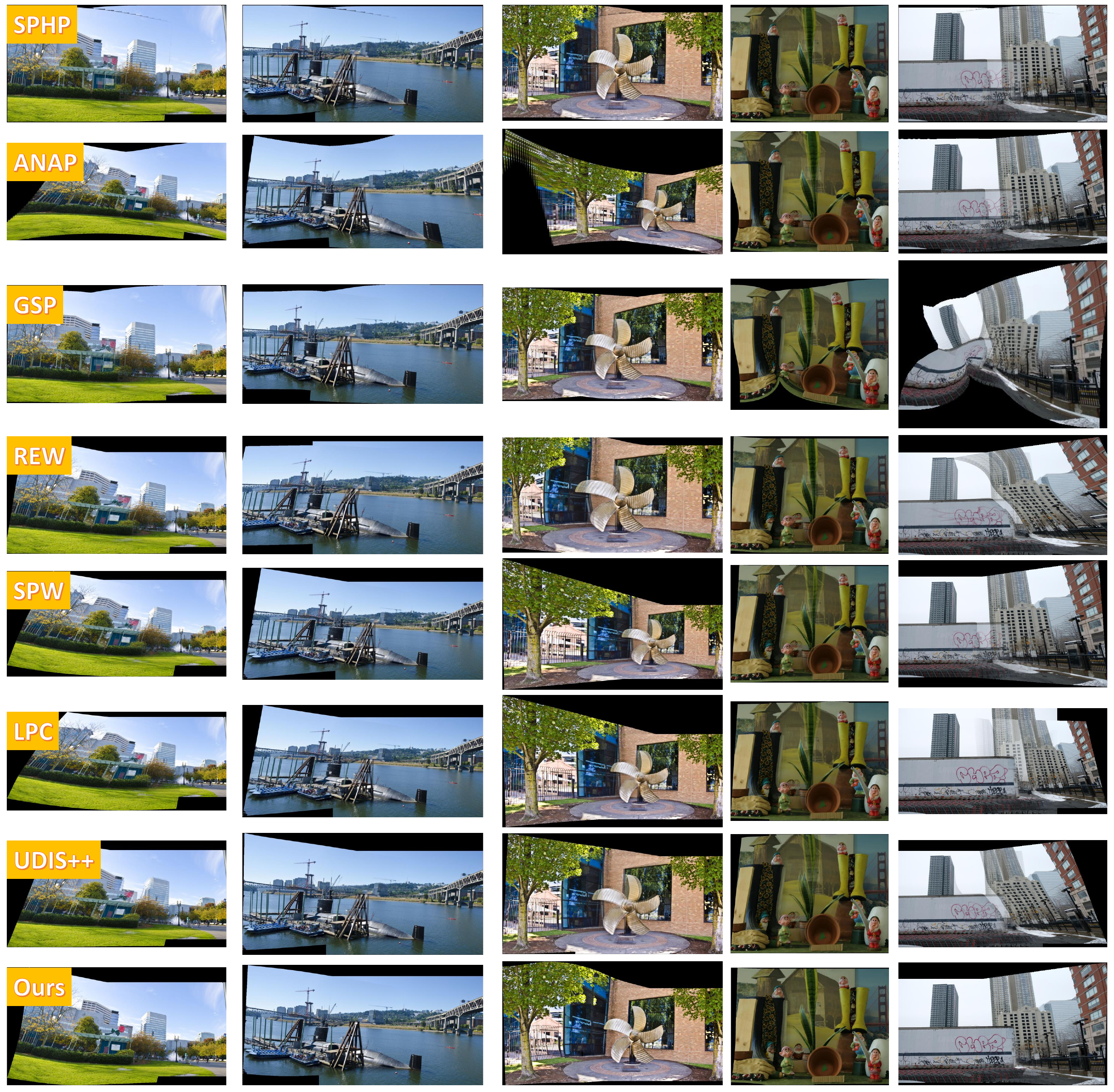}\\
	\caption{Comparisons of the image stitching results obtained by our method with that of the SOTA warping methods. Only part of the warping methods are included due to the limited space. (Best view in color and zoom in)}
	\label{fig:comp}
\end{figure*}

Then we compare the final visual panoramas with these SOTA warping methods in Fig.~\ref{fig:comp}, which contains five challenging image pairs. Large parallax occurs in the boundary between the foreground objects and background scenes. Existing warping methods indeed fail to align them due to the smoothness constraint, which results in undesirable ghosting artifacts in the blended panoramas. In contrast, with the utilization of the multiple homographies fitted on the segmented contents by SAM, our warping method aligns them simultaneously. Due to the limited space, we only show parts of the results. All the input image pairs and comparison results on the three datasets are provided in the supplementary material.

\subsection{Ablation Study}\label{sec:5-3}

\begin{figure*}[t]
	\centering
	\includegraphics[height=0.12\textheight]{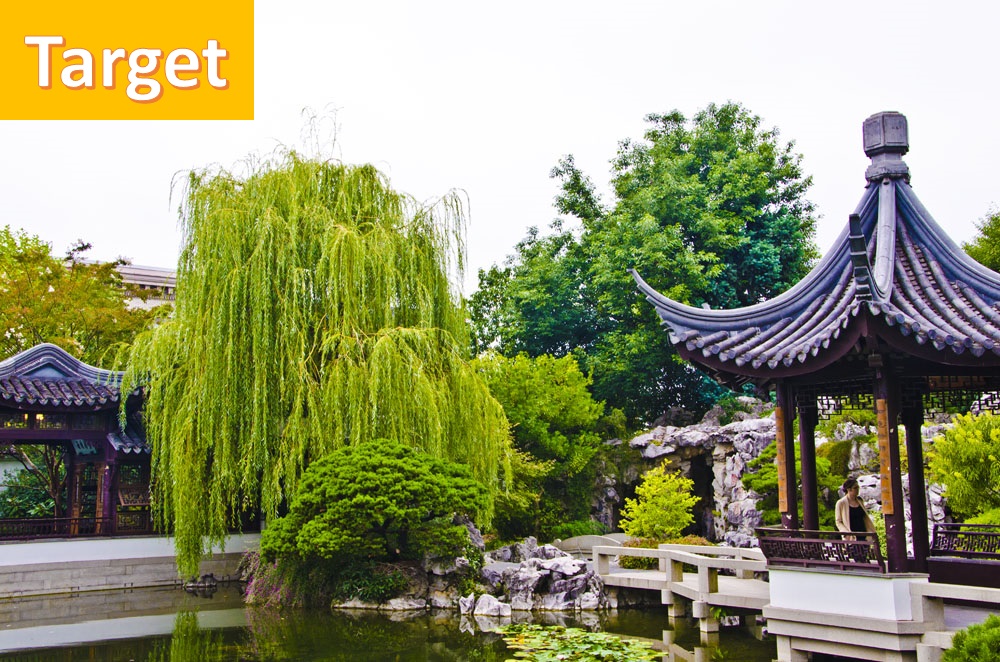}
	\includegraphics[height=0.12\textheight]{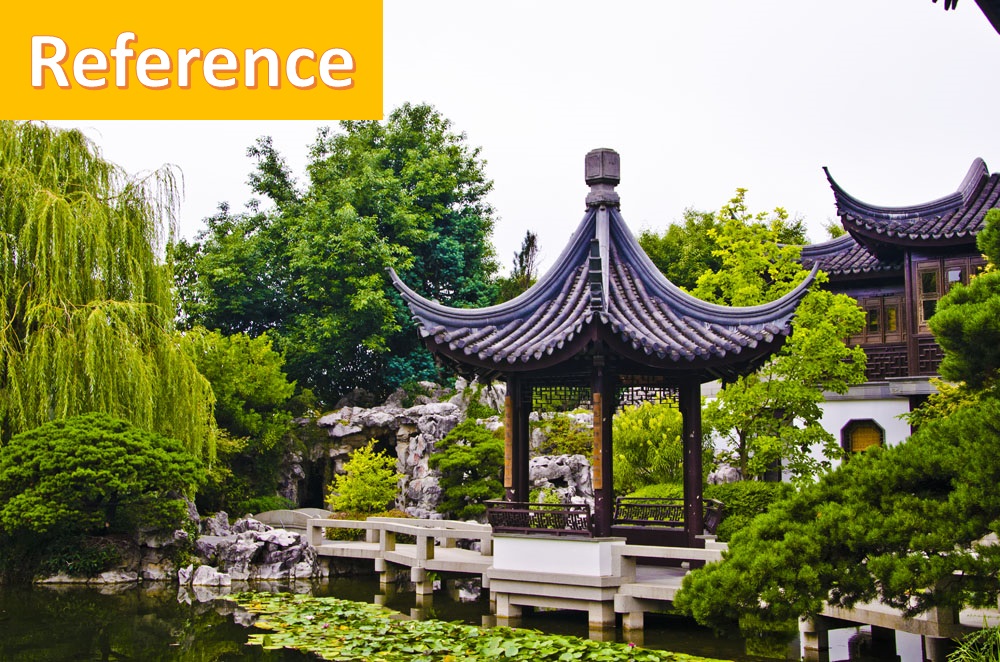}
	\includegraphics[height=0.12\textheight]{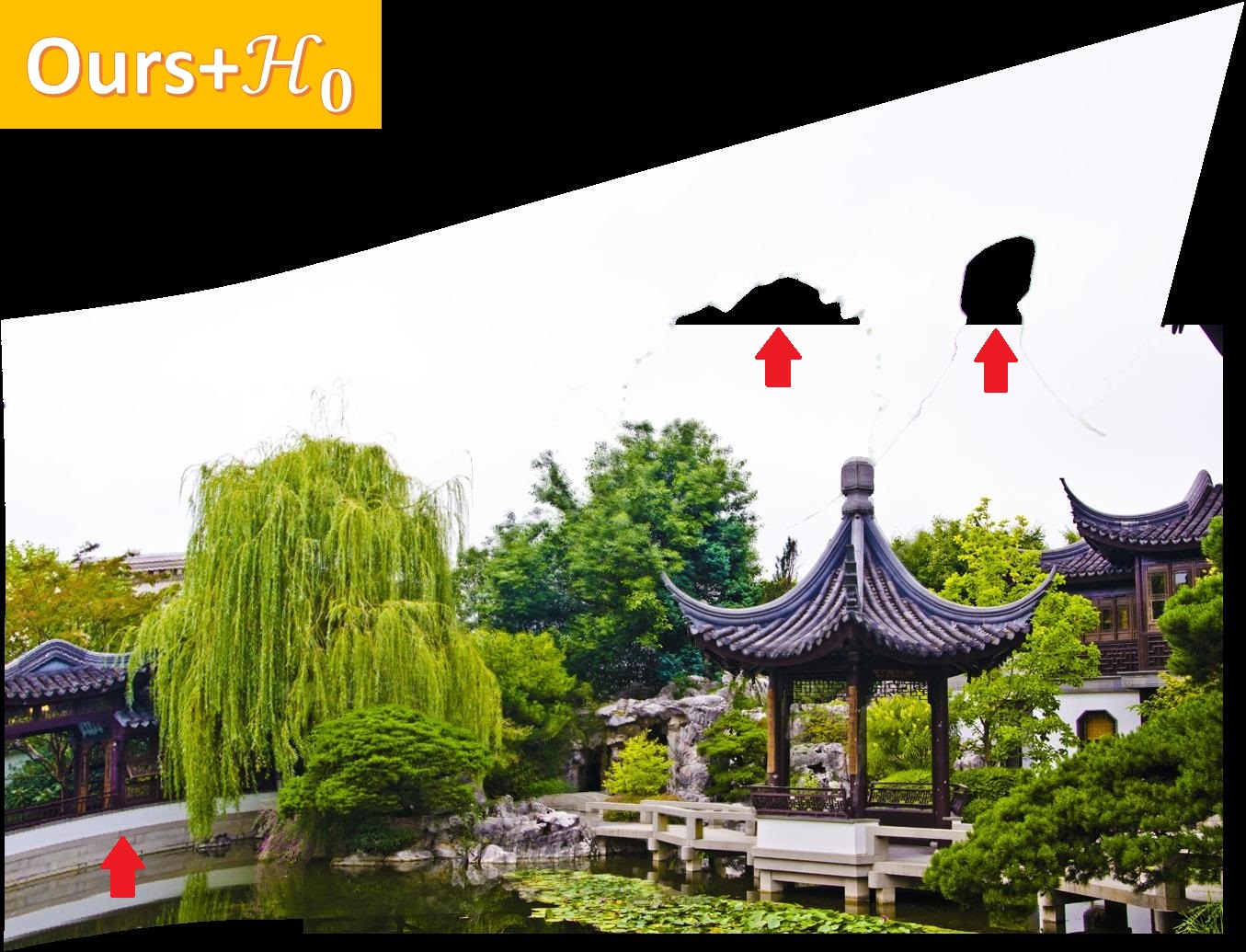}
	\includegraphics[height=0.12\textheight]{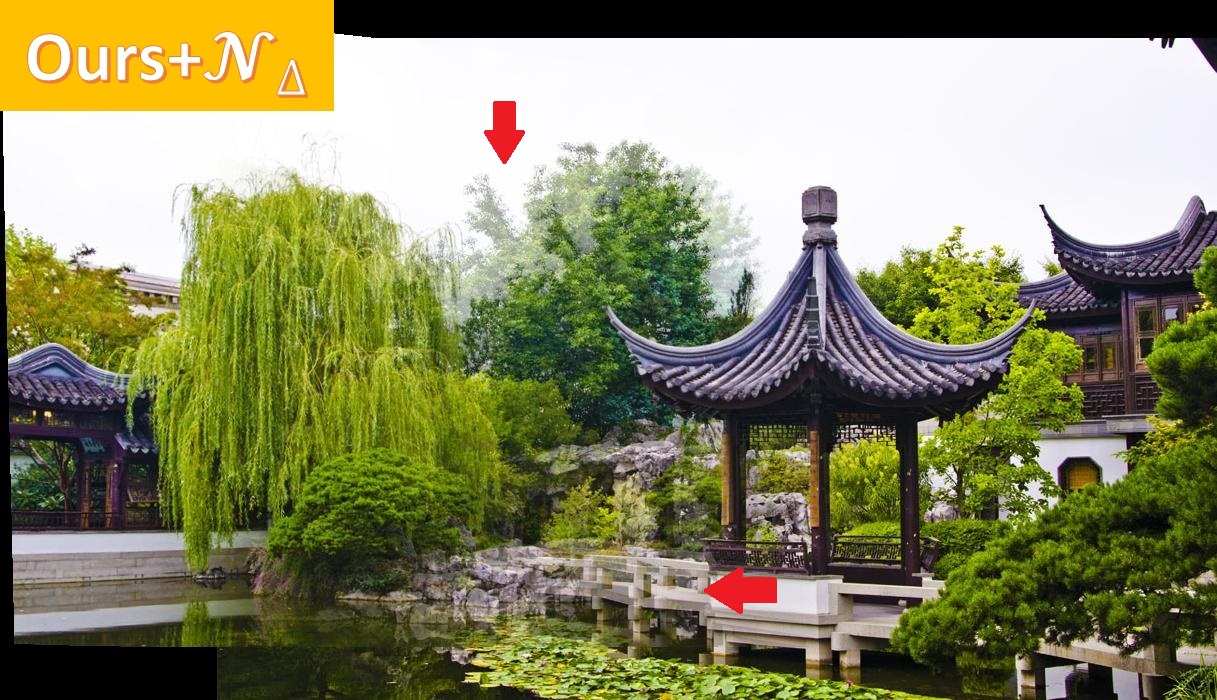}\\
	\includegraphics[height=0.138\textheight]{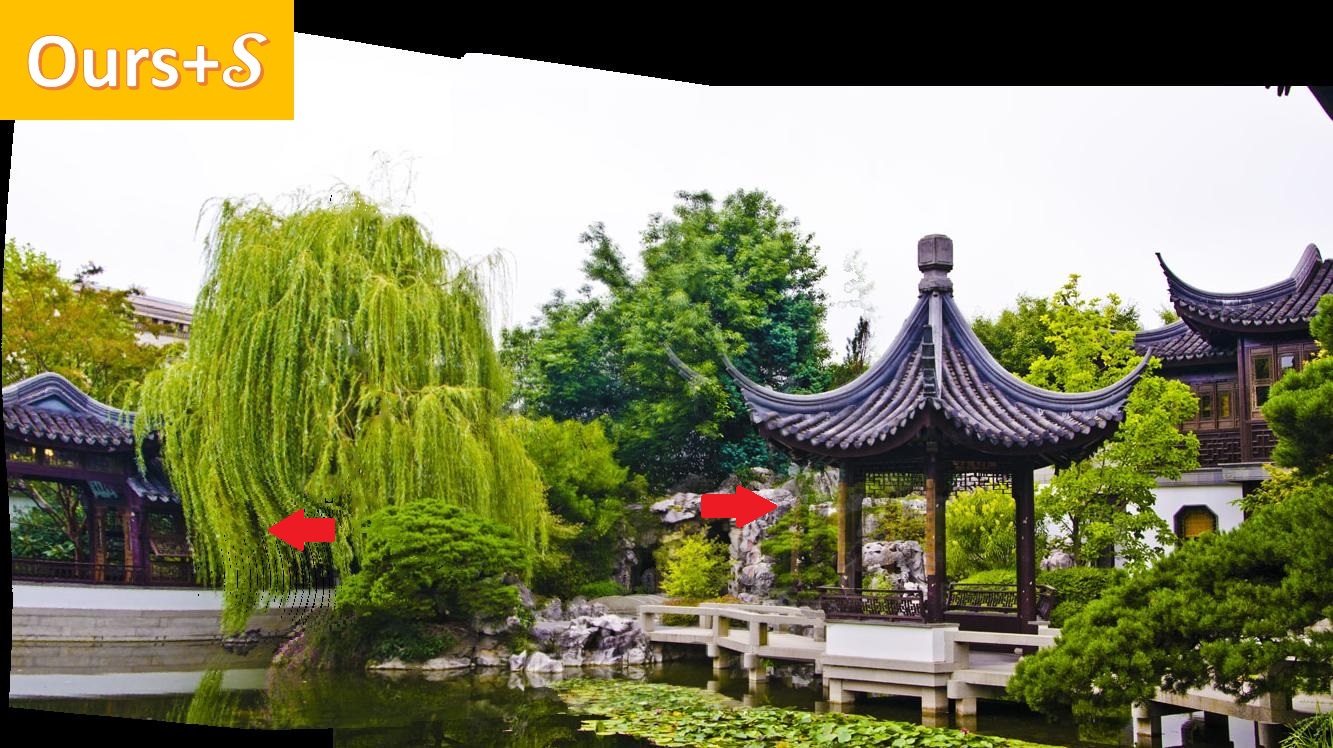}
	\includegraphics[height=0.138\textheight]{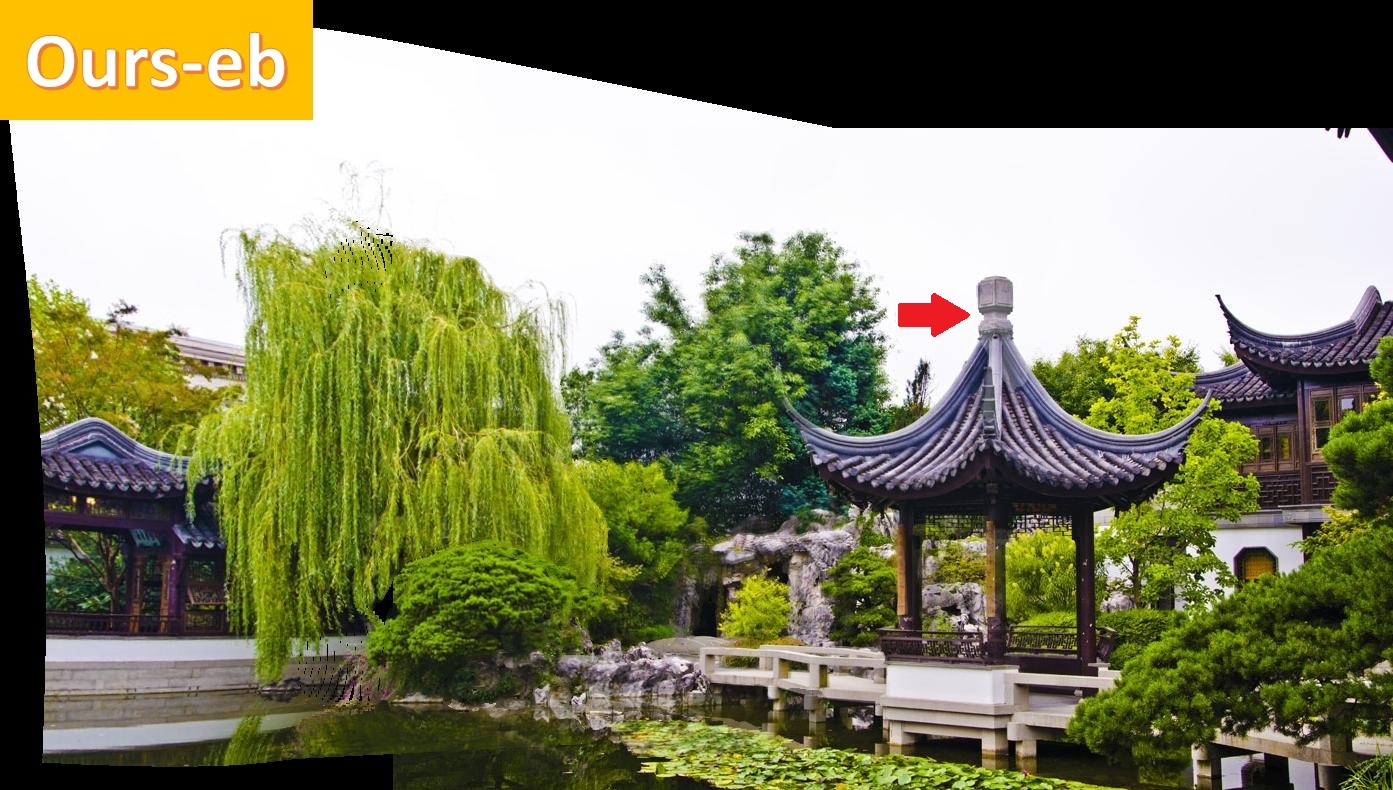}
	\includegraphics[height=0.138\textheight]{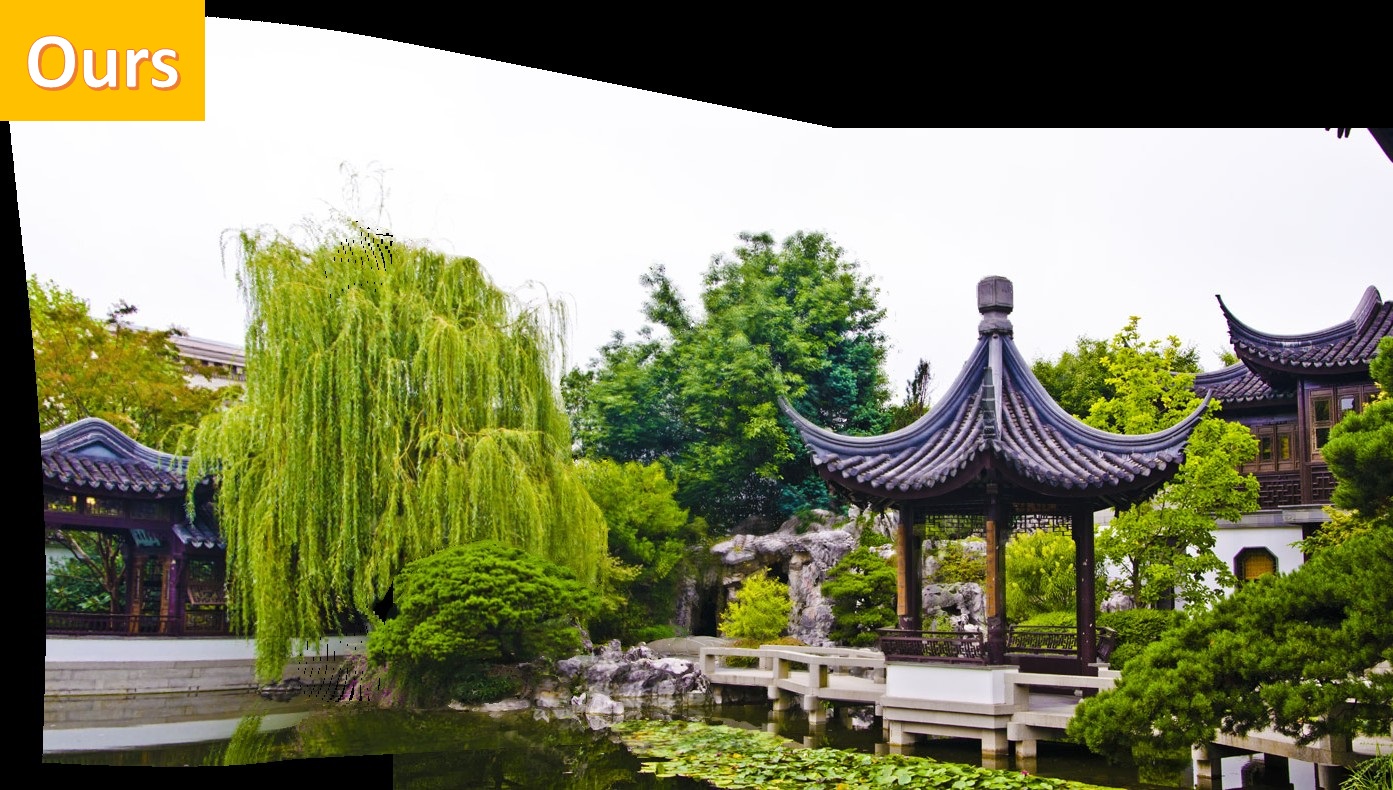}
	\caption{Visually comparisons between different ablation methods. Red arrows indicate the ghosting artifacts and distortions. The PSNR, SSIM and LPIPS metrics for ``Ours+$\mathcal{H}_0$'',  ``Ours+$\mathcal{N}_\triangle$'', ``Ours+$\mathcal{S}$'', ``Ours-\textbf{eb}'' and ``Ours'' are (\textbf{18.09}, 11.65, 16.50, 16.86, \underline{17.76}), (\textbf{0.779}, 0.493, 0.701, 0.742, \underline{0.756}), (\underline{0.178}, 0.432, 0.230, 0.202, \textbf{0.177}). (Best view in color and zoom in).}
	\label{fig:ablation}
\end{figure*}

To validate the effectiveness of each module in our method, We conduct the ablation study on the three datasets and demonstrate the results in Table~\ref{tab:ablation}. Among them, ``Ours+$\mathcal{H}_0$'' is proposed to ablate the module of the multi-homography fitting step, and the initial homography models $\mathcal{H}_0$ are instead introduced here to label and align the segmented contents; ``Ours+$\mathcal{N}_\triangle$'' denotes that we only use Delaunay triangulation to define the neighborhood system in the multi-homography fitting; and ``Ours+$\mathcal{S}$'' represents that we replace SAM by superpixel segmentation \cite{Achanta2012SLIC} to align different image contents; ``Ours-\textbf{eb}'' means that we do not use the error-buffer algorithm in the image warping. Experiments show that including SAM to define the neighborhood system and using our error-buffer algorithm improves the alignment performance, benefiting from the accurate semantic information. Despite that ``Ours+$\mathcal{H}_0$'' slightly improves some metrics in the three datasets, our final model is more stable and visually pleasing. The first three ablated methods all have several failure cases when conducted on the entire datasets. We demonstrate one visual comparison result in Fig. \ref{fig:ablation}, using $\mathcal{H}_0$ may generate unnatural distortions in the non-overlapping region, using $\mathcal{S}$ (superpixel) may have ghosting artifacts due to the inaccurate segmentation on the object boundary.

\begin{table*}
	%\scriptsize
	\begin{center}
		\caption{Ablation study. The ablation methods have failing image pairs that are not included in the evaluation. The best is marked in bold and the second best is in underline.} \label{tab:ablation}%
		\begin{tabular}{l|ccc|ccc|ccc|c}
			\hline
			& \multicolumn{3}{c|}{Parallax \cite{zhang2014parallax}} & \multicolumn{3}{c|}{SEAGULL \cite{lin2016seagull}} & \multicolumn{3}{c|}{MR \cite{herrmann2018robust}}\\
			Method   & PSNR $\uparrow$ & SSIM $\uparrow$  & LPIPS $\downarrow$ & PSNR $\uparrow$  & SSIM $\uparrow$  & LPIPS $\downarrow$ & PSNR $\uparrow$  & SSIM $\uparrow$  & LPIPS $\downarrow$ & \# FC\\
			\hline
			Ours+$\mathcal{H}_0$ & \textbf{19.24} & \textbf{0.755} & \underline{0.199} & \textbf{18.99} & \textbf{0.716} & \underline{0.227} & \underline{18.36} & \textbf{0.741} & 0.204 & 3\\
			Ours+$\mathcal{N}_\triangle$ & 17.44 & 0.696 & 0.227 & 17.84 & 0.670 & 0.244 & 15.96 & 0.633 & 0.269 & 4\\
			Ours+$\mathcal{S}$ & 18.50  & 0.730 & 0.211 & 18.02 & 0.697 & 0.244 & 17.27 & 0.682 & 0.244 & 3\\
			%			Ours+$\mathcal{S}100$ & \textbf{19.51}  & \textbf{0.769} & \textbf{0.187} & 18.41 & 0.701 & 0.235 & 17.90 & 0.710 & 0.223 & 3\\
			%			Ours+$\mathcal{S}150$ & \underline{19.50}  & \underline{0.759} & 0.194 & \underline{18.77} & 0.711 & 0.252 & 18.07 & 0.715 & 0.221 & 2\\
			Ours$-$\textbf{eb} & 18.75  & 0.746 & 0.202 & 18.29 & 0.702 & 0.240 & 18.11 & 0.732 & \underline{0.202} & 0\\
			Ours & \underline{19.19} & \underline{0.752}  & \textbf{0.192} & \underline{18.69} &  \underline{0.713} & \textbf{0.226} & \textbf{18.55} & \underline{0.736} & \textbf{0.197} & 0\\
			\hline % 1,1,1; 2,1,1; 2,2,0 
		\end{tabular}%
	\end{center}
\end{table*}%

\begin{figure*}[t]
	\centering
	\subfloat[]{
		\includegraphics[height=0.167\textheight]{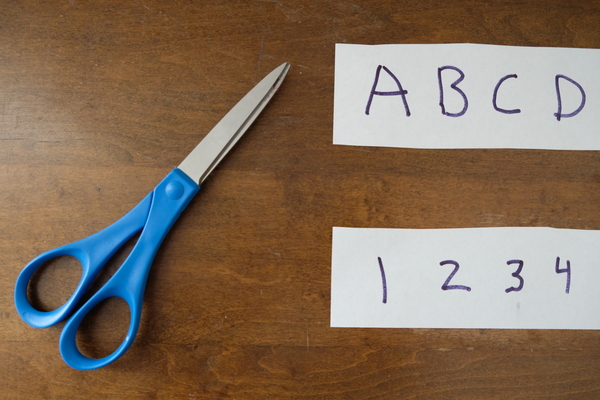}
		\includegraphics[height=0.167\textheight]{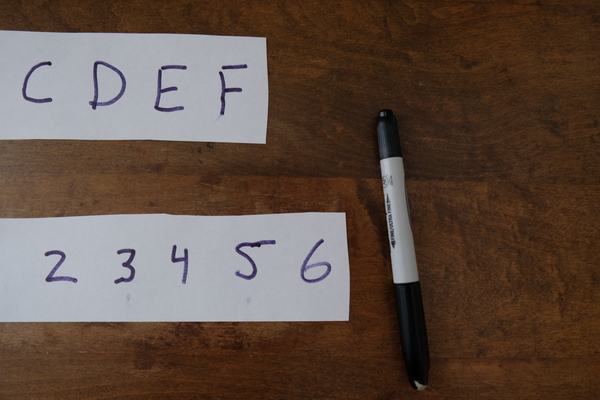}}
	\subfloat[]{
		\includegraphics[height=0.167\textheight]{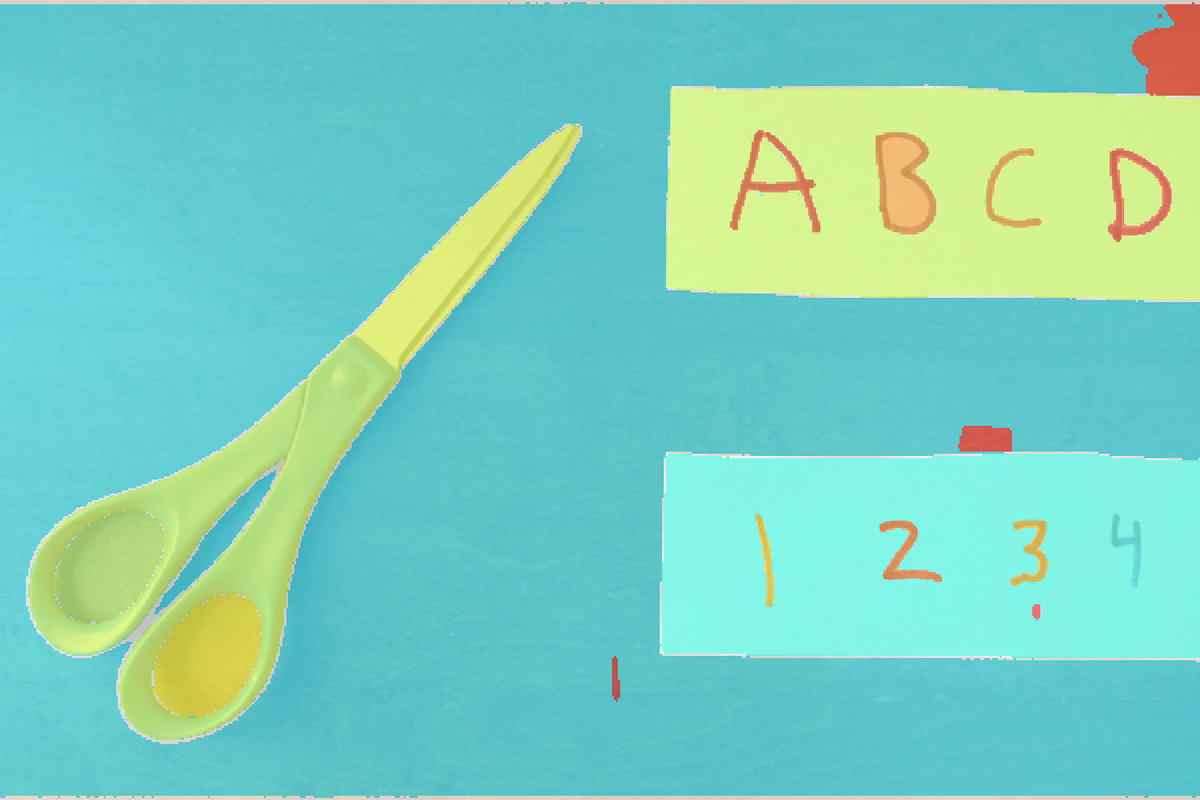}}\\
	\subfloat[]{
		\includegraphics[height=0.14\textheight]{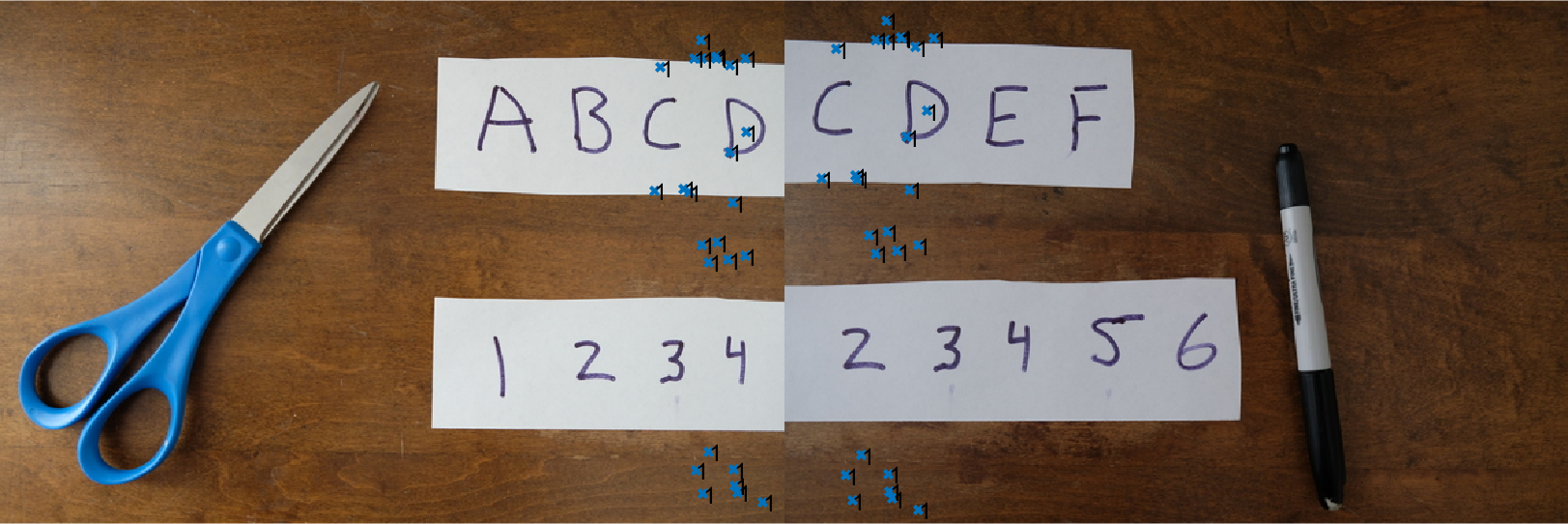}}
	\subfloat[]{
		\includegraphics[height=0.14\textheight]{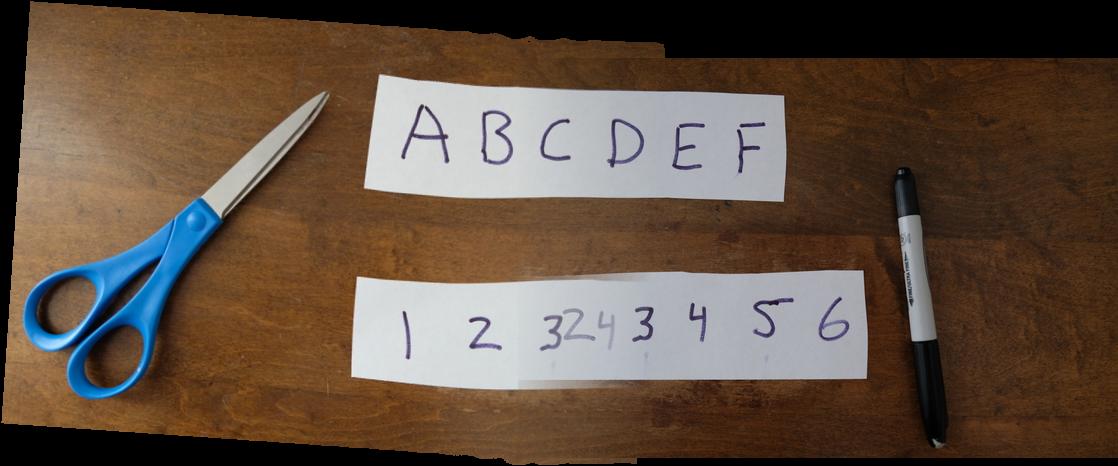}}\\
	\caption{Failure example of our method. (a) Input images. (b) The SAM result of the target image. (c) Feature matches after our multi-homography fitting algorithm. (d) Our final result. (Best view in color and zoom in)}
	\label{fig:failure}
\end{figure*}

\subsection{Discussion}

%{\color{blue}\textbf{Discussion.}} 
Our warping method calculates multiple homographies via the subsets of feature points. Note that the total number of feature points should not be too small, otherwise, the fitted homography may fail to align the segmented contents. Additionally, the homography linearization step only constrains $\mathcal{C}^1$ smoothness of the extrapolation from overlapping to non-overlapping region. Fig. \ref{fig:failure} shows a failure example of our method. It lacks a sufficient number of feature matches between the input images, such that our method only generates one single homography and the final result has severe misalignments. Using SAM to help construct region matches and line feature matches can improve the alignment accuracy. Besides, adopting higher continuity and other structure-preserving constraints will enable the warping results more visually pleasing. We leave these as the future work.

\section{Conclusions}
\label{sec:conclusion}

In this paper, we propose a multi-homography warping method for handling images with large parallax. We first segment the target image into semantic contents via SAM, and then partition the feature points into multiple subsets via the multi-homography fitting algorithm, where each subset fits a homography and aligns the part of the image contents well. Subsequently, we label each semantic content in the overlapping region using the best-fitting homography with the lowest photometric error. The best-fitting homographies are then linearized to the non-overlapping region to constrain a smooth warp. Experimental results demonstrate that our proposed method accurately aligns challenging images with large parallax, and yields a significantly better performance compared with the state-of-the-art image warping methods.
%\clearpage
{
    \small
    \bibliographystyle{ieeenat_fullname}
    \bibliography{main}
}

% WARNING: do not forget to delete the supplementary pages from your submission 
% \input{sec/X_suppl}

\end{document}